\ifdefined\pdfobjcompresslevel
\fi
\PassOptionsToPackage{table,dvipsnames}{xcolor}
\documentclass[]{selfevolagent}
\usepackage{microtype}
\usepackage{amsfonts}
\usepackage{amsmath,amssymb,mathtools}
\usepackage{xcolor}
\usepackage{wrapfig}
\usepackage{float}
\usepackage{array}
\usepackage{colortbl}
\usepackage{makecell}
\usepackage{multirow}
\usepackage{tabularx}
\usepackage{enumitem}
\usepackage{pifont}
\usepackage{tikz}
\usepackage{graphicx}
\usepackage{booktabs}
\usepackage[most]{tcolorbox}
\usepackage{url}
\usepackage{fontawesome5}
\usepackage{caption}
\usepackage{subcaption}
\usepackage{titletoc}
\ifPDFTeX
\usepackage[T1]{fontenc}
\fi

\definecolor{oursrow}{HTML}{EDE7F6}
\definecolor{reflexred}{HTML}{A51C30}
\definecolor{skillblue}{HTML}{2B6CB0}
\definecolor{recovergreen}{HTML}{2F855A}
\definecolor{noiseorange}{HTML}{C05621}
\definecolor{softgray}{HTML}{F3F4F6}
\definecolor{skillpurple}{HTML}{6B4AA3}
\definecolor{grouprow}{HTML}{FAF1D8}
\definecolor{capsuleblue}{HTML}{1B4F72}
\definecolor{capsulelight}{HTML}{EAF3F8}
\definecolor{blockblue}{HTML}{D6EAF8}
\definecolor{blockred}{HTML}{FDEDEC}
\definecolor{blockgreen}{HTML}{E8F6F3}

\definecolor{AppendixTOCSection}{HTML}{4B2E83}
\definecolor{AppendixTOCSubsection}{HTML}{6B4AA3}
\definecolor{AppendixTOCSubsubsection}{HTML}{8A74B8}
\newcommand{\setupappendixtoc}{%
    \titlecontents{section}
        [2.6em]
        {\addvspace{0.35em}\small\bfseries\color{AppendixTOCSection}}
        {\contentslabel{2.3em}}
        {}
        {\titlerule*[0.45pc]{.}\contentspage}
    \titlecontents{subsection}
        [3.9em]
        {\small\color{AppendixTOCSubsection}}
        {\contentslabel{2.8em}}
        {}
        {\titlerule*[0.45pc]{.}\contentspage}
    \titlecontents{subsubsection}
        [6.2em]
        {\footnotesize\color{AppendixTOCSubsubsection}}
        {\contentslabel{3.5em}}
        {}
        {\titlerule*[0.45pc]{.}\contentspage}%
}

\newtcolorbox{skillcapsule}[1]{%
  enhanced, breakable,
  colback=capsulelight, colframe=capsuleblue, coltitle=white,
  boxrule=0.6pt, arc=3pt,
  left=8pt, right=8pt, top=4pt, bottom=6pt,
  before skip=8pt, after skip=10pt,
  fonttitle=\large\sffamily\bfseries, fontupper=\small,
  title={#1}%
}
\newcommand{\skillblock}[3]{%
  \par\smallskip
  \noindent\colorbox{#1}{\textbf{#2}}
  \par\smallskip #3\par%
}

\newcommand{\method}{\textsc{Agent-Reflex}}
\newcommand{\rsd}{\textsc{RSD}}
\newcommand{\resft}{\textsc{ReSFT}}
\newcommand{\skill}{\kappa}

\title{Agent-Reflex:\\Teaching Language Agents to Act Reflectively}

\author[1]{Guanting Dong}
\author[1]{Xinyu Yang}
\author[1]{Jiajie Jin}
\author[1]{Yuyang Hu}
\author[1]{Xiaoshuai Song}
\author[1]{Yifei Chen}
\author[1]{Xiaoxi Li}
\author[1]{Jiejun Tan}
\author[1]{Yutao Zhu}
\author[2]{Tat-Seng Chua}
\author[1]{Ji-Rong Wen}
\author[1,\dagger]{Zhicheng Dou}

\affiliation[1]{Gaoling School of Artificial Intelligence, Renmin University of China}
\affiliation[2]{National University of Singapore}
\contribution[\dagger]{Corresponding author}

\abstract{%
Large language models have evolved from single-turn chatbots into interactive agents capable of long-horizon reasoning. In real deployments, MCP tool services can return noisy feedback and users may intervene to correct the agent, but existing methods provide limited supervision for reflection learning and failure recovery. To bridge this gap, we propose \method{}, a general training framework for incentivizing proactive reflection in agents. Specifically, \method{} comprises three stages. First, Environment Reflection Skill Discovery (\rsd{}) injects grounded tool-feedback perturbations into environment scaling, samples a capability-diverse LLM pool, mines matched success--failure pairs, and summarizes reusable reflection meta-skills. Building on this skill library, we then apply Reflective Supervised Fine-Tuning (\resft{}) that converts each skill into supervision for self-reflection, skill recovery, and user-centric correction, strengthening complementary reflective behaviors. Finally, we propose On-Policy Reflective Policy Optimization (ORPO) to combine agentic reinforcement learning under noisy interactions with reflection-guided on-policy distillation to further elicit proactive reflection. Built on Qwen3-8B, \method{} reaches 58.6 overall on five long-horizon tool-use benchmarks and 29.1 on AgentNoiseBench, surpassing strong environment-scaling and reflection baselines. Experiments on six challenging agent benchmark suites demonstrate the effectiveness of \method{}. Further analyses show that \method{} improves agents' robustness to noise while maintaining stable training dynamics, supporting scalable agent training.
}
\metadata[\faEnvelope\ Contact]{dongguanting@ruc.edu.cn, dou@ruc.edu.cn}
\metadata[{\raisebox{-0.2ex}{\includegraphics[height=1.1em]{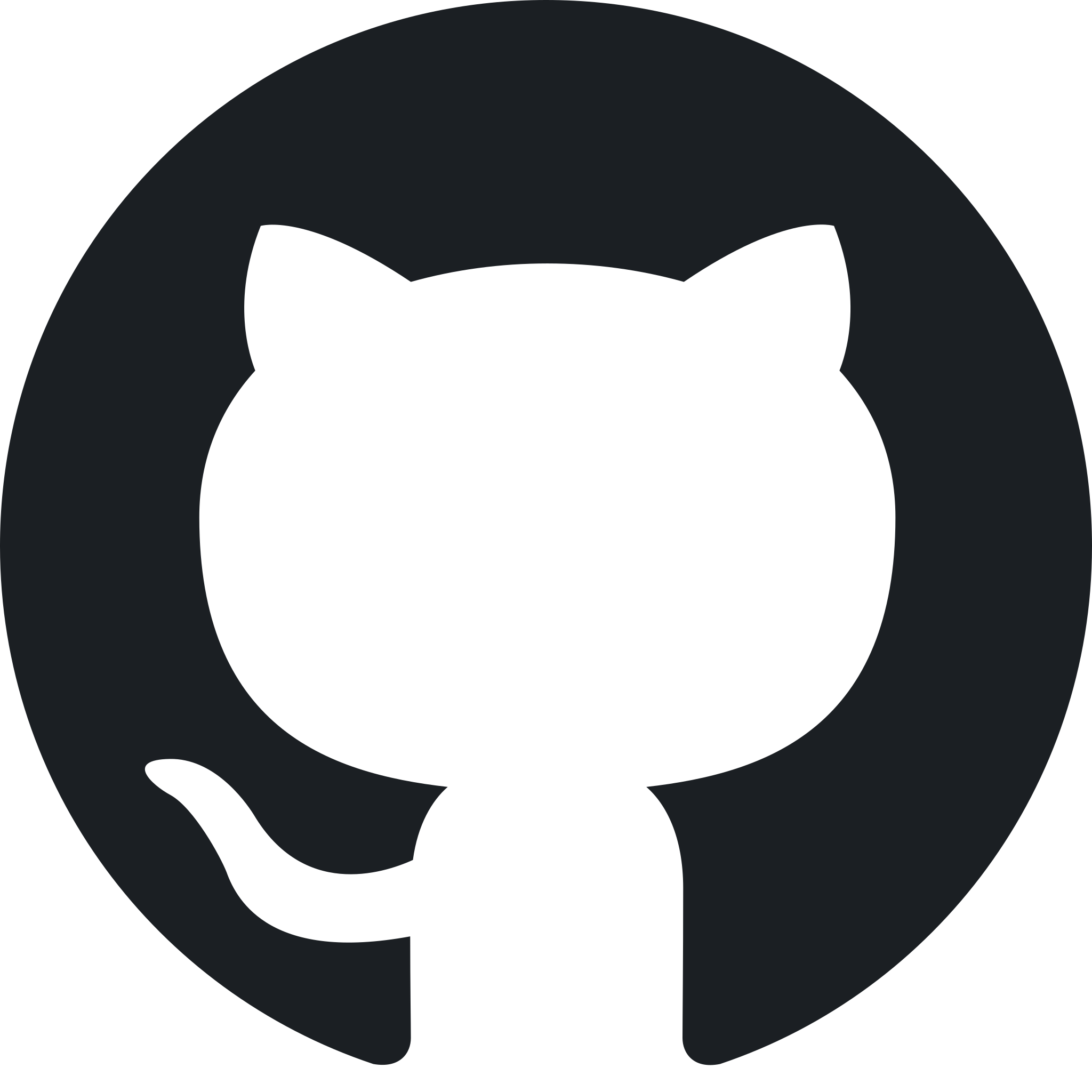}}\ Code}]{\url{https://github.com/RUC-NLPIR/Agent-Reflex}}
\metadata[{\raisebox{-0.25ex}{\includegraphics[height=1.15em]{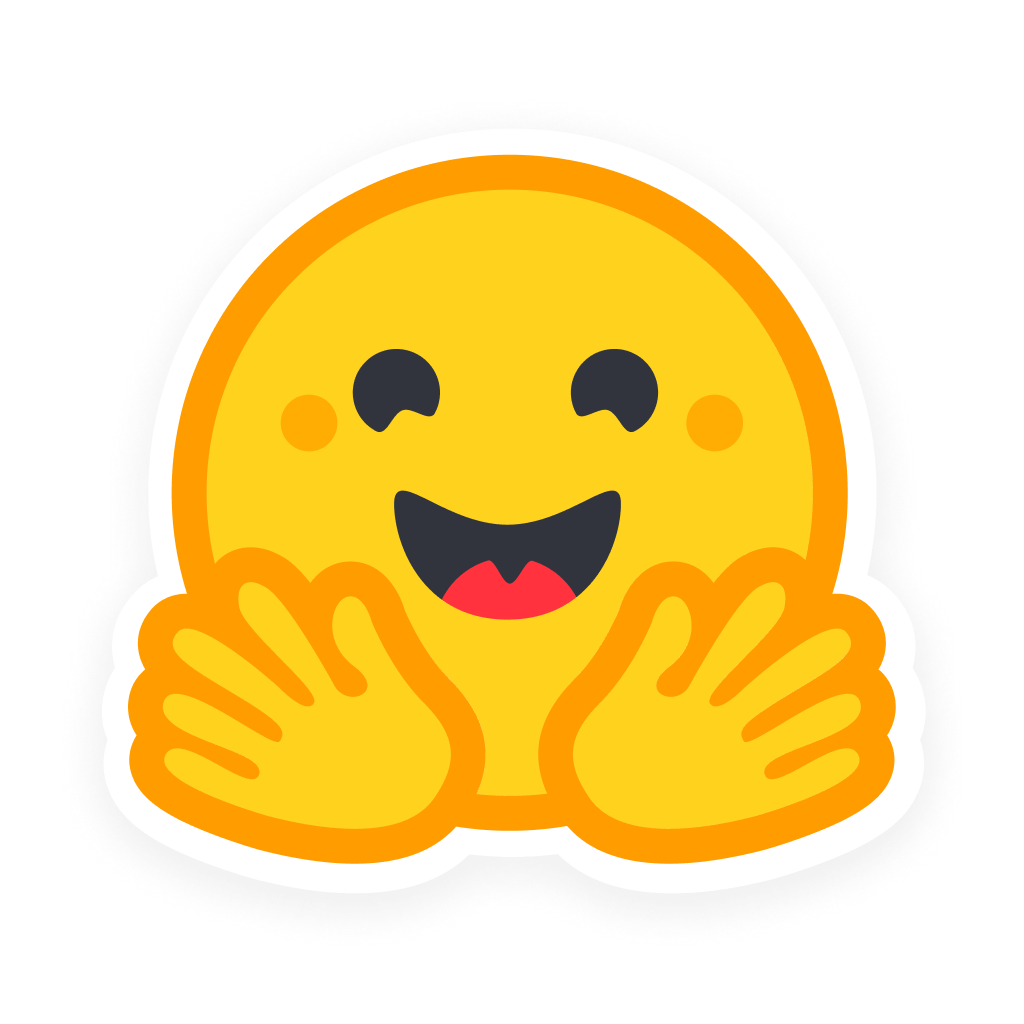}}\ Models \& Data}]{\url{https://huggingface.co/collections/dongguanting/agent-reflex}}

\begin{document}

\maketitle

\vspace*{0.3cm}
\begin{figure}[H]
  \centering
  \includegraphics[width=\linewidth]{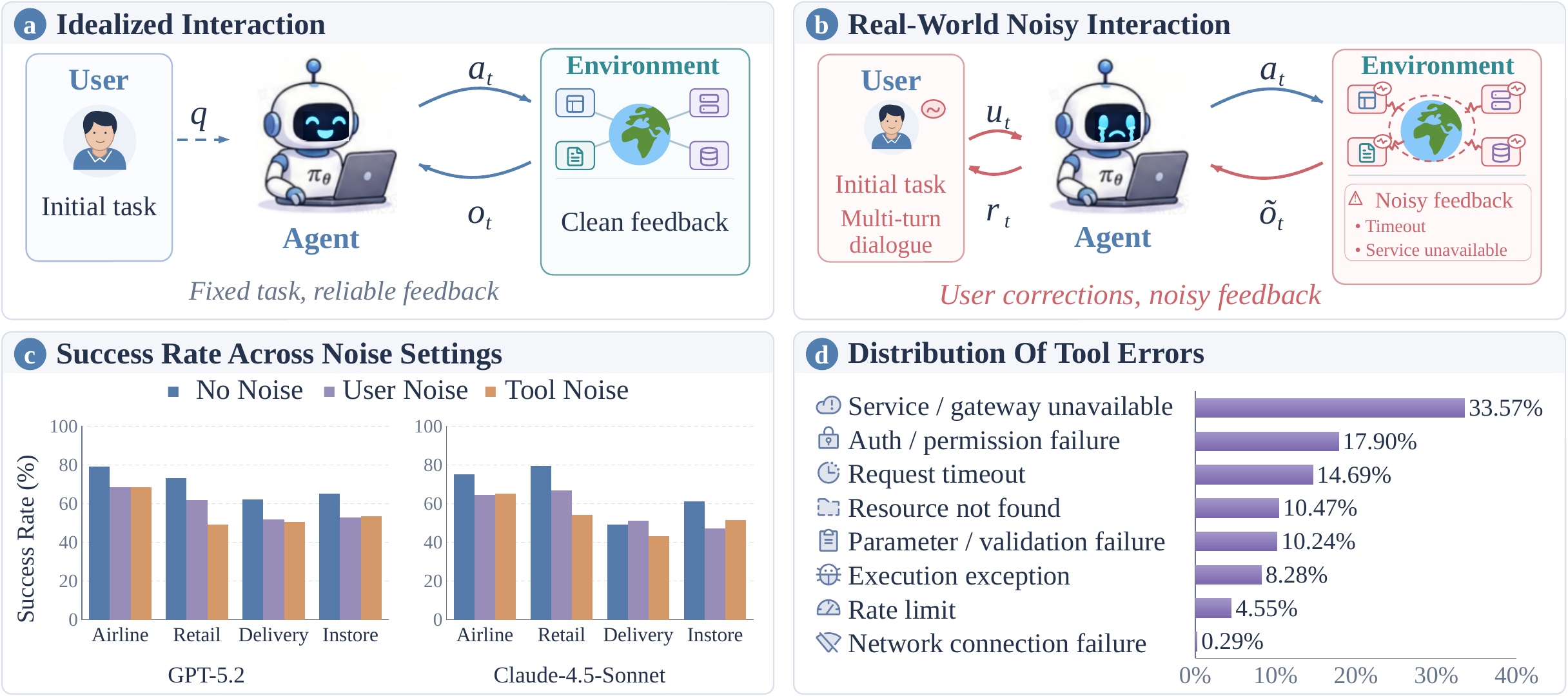}
  \caption{\textbf{From idealized interaction to reflective recovery.} \textbf{(a)} Prior environments assume clean tool feedback, whereas \textbf{(b)} real ones return noisy observations and user corrections at arbitrary turns. \textbf{(c)} Success rate of GPT-5.2 and Claude-4.5-Sonnet under no, user, and tool noise. \textbf{(d)} Share of the eight tool-error families mined from 24,510 real error returns.}
  \label{fig:motivation}
\end{figure}

\clearpage
\section{Introduction}
Recent advances in large language models (LLMs) have extended their applications from natural language tasks to practical workflows that require long-horizon reasoning~\citep{yao2023react,patil2025bfcl,trivedi2024appworld,barres2025tau2,dong2026longhorizon}. Meeting this demand requires agents not only to master tool interfaces but also to accumulate transferable interaction experience from dynamic environments. Building diverse, stateful environments that faithfully capture real-world interaction logic has therefore become the foundation for scaling agent capability.

Recent studies to scale agent training environments can be roughly categorized into two groups. One group, exemplified by Simia~\citep{li2025simia} and GenEnv~\citep{guo2025genenv}, treats LLMs as implicit textual world models that generate feedback from tool specifications and interaction histories. These approaches substantially reduce environmental engineering costs, but the feedback generated can be hallucinated. Instead, a complementary family explicitly reproduces real interaction logic, synthesizing executable tools and environment databases: it extracts databases from real-world service themes, generates executable tools, and synthesizes tasks with verifiable results~\citep{song2026envscaler,wang2026awm,tu2026scaleenv,dong2026agentworld}. As a result, these executable environments give agents a basis for learning interaction patterns that transfer across environments.

Although environment scaling improves interaction diversity and out-of-distribution generalization, existing agent-training paradigms still rely on an idealized assumption: agents interact with stable, well-controlled environments under minimal user intervention. This mismatch with real-world deployment reveals two inherent challenges (as shown in Figure~\ref{fig:motivation}):

\textbf{(1) Lack of noisy feedback.} Real-world tool services are unreliable, frequently returning timeouts and service errors. Existing training environments usually assume stable and noise-free feedback, leaving agents underprepared for failure detection and recovery~\citep{wang2026agentnoisebench,gupta2026reliabilitybench}.

\textbf{(2) Lack of user-guided correction.} Real users actively intervene when an agent deviates from their intent, correcting its reasoning and actions. Existing user simulators often rely on standardized templates, with limited coverage of corrective instructions and diverse interaction styles during execution~\citep{barres2025tau2,qian2025userrl,suhr2023realtime}.

Together, these challenges reveal that agents still lack proactive reflection under noisy external feedback. Ideally, a robust agent should continuously perceive environmental noise, task failures, and user correction intent, then promptly revise its actions. Existing reflection methods explore inference-time reflection, correction-trajectory learning, and noise-augmented training~\citep{shinn2023reflexion,madaan2023selfrefine,yuan2025agentr,bi2026reflectrl,zhang2026loongreflect,chen2026noisyagent}, but most are trained in stateless or simplified environments, leaving one question open: \textbf{\emph{How can we elicit proactive reflective capability in agents from execution failures in real-world environments?}}

To address this gap, we propose \method{}, a general training framework for incentivizing proactive reflective thinking in agents. It models reflection as a learnable meta-skill: transferable meta-skills are explicitly mined from execution failures and progressively internalized into the policy through verifiable reflection learning. Specifically, \method{} comprises three components:

\textbf{(1) Environment Reflection Skill Discovery (RSD).} We anchor 557 high-usage real-world environment themes and automatically construct database-backed executable environments and tasks, injecting diverse tool-return failures to emulate noisy conditions. From a diverse LLM agent pool, we sample trajectories covering varied task types and response styles, and matched success/failure pairs are then mined into 1,170 reflection meta-skills across 557 environments.

\textbf{(2) Reflective Supervised Fine-Tuning (ReSFT).} Given the skill library, we design three auxiliary and verifiable reflection sampling strategies: \textit{Self-Reflection} (SR), \textit{Skill Recovery} (KR), and \textit{User-Centric Reflection} (UR), which strengthen reflection at different levels. Supervised fine-tuning then internalizes these explicit skills into the agent policy.

\textbf{(3) On-Policy Reflective Policy Optimization (ORPO).} To further elicit proactive reflection, ORPO combines Noisy Agent RL with reflection-guided on-policy distillation. For each task we match its environment's reflection meta-skill and rollout in the noisy environment, optimizing the policy by reinforcement learning from verifiable rewards. The same action tokens are then re-scored under a privileged skill context, and a confidence-weighted OPD objective~\citep{wu2026seed} turns this contrast into dense supervision for reflection decisions.

Experiments on six long-horizon tool-use and noisy-interaction benchmarks demonstrate the effectiveness of \method{}. Further analysis shows that it improves noise robustness and keeps training dynamics stable, supporting scalable general-agent training.

Therefore, our contributions are threefold:
\begin{itemize}[leftmargin=1.25em,itemsep=1pt,topsep=2pt]
  \item We identify two gaps between idealized synthetic environments and real-world
  deployment in agent interaction, namely \textit{lack of noisy environment feedback} and \textit{lack of user-guided correction
  during execution}, highlighting the importance of reflection learning for noisy interactions.
  \item We introduce \rsd{}, which systematically builds a scalable noisy-environment ecosystem by
  incorporating real-tool perturbations and mining reflection meta-skills from large-scale success/
  failure trajectory pairs sampled by an LLM Agent Pool.
  \item We further propose \resft{} and ORPO. \resft{} builds multi-level reflection data through
  three sampling strategies and internalizes it by supervised fine-tuning; ORPO adds
  reflection-guided on-policy distillation to Noisy Agent RL for token-level supervision of failed
  executions.  
\end{itemize}

\section{Related Work}
\label{sec:related-work}

\textbf{Agent Reflection Learning.} Agent reflection differs in where the corrective signal lives. Inference-time methods leave parameters untouched: Reflexion and Self-Refine append a verbal self-critique, and ExpeL retrieves cross-task insights at test time~\citep{shinn2023reflexion,madaan2023selfrefine,zhao2024expel}. Because the critique comes from the policy that failed, a well-formed but wrong tool return is rarely contradicted. Training-time methods move the signal into a per-trajectory repair: Agent-R, ReflectRL, and LoongReflect learn from that repair~\citep{yuan2025agentr,bi2026reflectrl,zhang2026loongreflect}, and NoisyAgent exposes the policy to perturbed interaction under only a scalar return~\citep{chen2026noisyagent}. Either way, nothing accumulates into an environment-level statement a later episode can reuse. \method{} mines cross-model success/failure contrasts under the same injected noise into environment-indexed meta-skills and internalizes them through two-stage training.

\textbf{Scalable Environment Synthesis for Agent Training.} Diverse, stateful environments underpin tool use and long-horizon decision making. One line of work simulates interaction feedback with language models, treating tool specifications and dialogue history as an implicit world model~\citep{li2025simia,guo2025genenv}. A second makes the environment itself executable: EnvScaler and ScaleEnv synthesize tools with verifiable tasks~\citep{song2026envscaler,tu2026scaleenv}, Agent-World scales database-backed services over hundreds of real-world themes and verifies outcomes through state transitions~\citep{dong2026agentworld,wang2026awm}, and EnvACE internalizes environment dynamics through world rehearsal~\citep{xu2026envace}. All of them scale the \emph{supply} of solvable environments, whereas \method{} keeps the database-backed paradigm but turns the failures inside those environments into supervision, mining diagnosis and recovery rules from cross-model success/failure contrasts under perturbations derived from real tool errors. Appendix~\ref{app:related-work} gives an extended discussion of reflection learning, environment synthesis, and agentic RL.

\section{Methodology: Agent-Reflex}
\label{sec:method}
\begin{figure}[t]
  \centering
  \includegraphics[width=\linewidth]{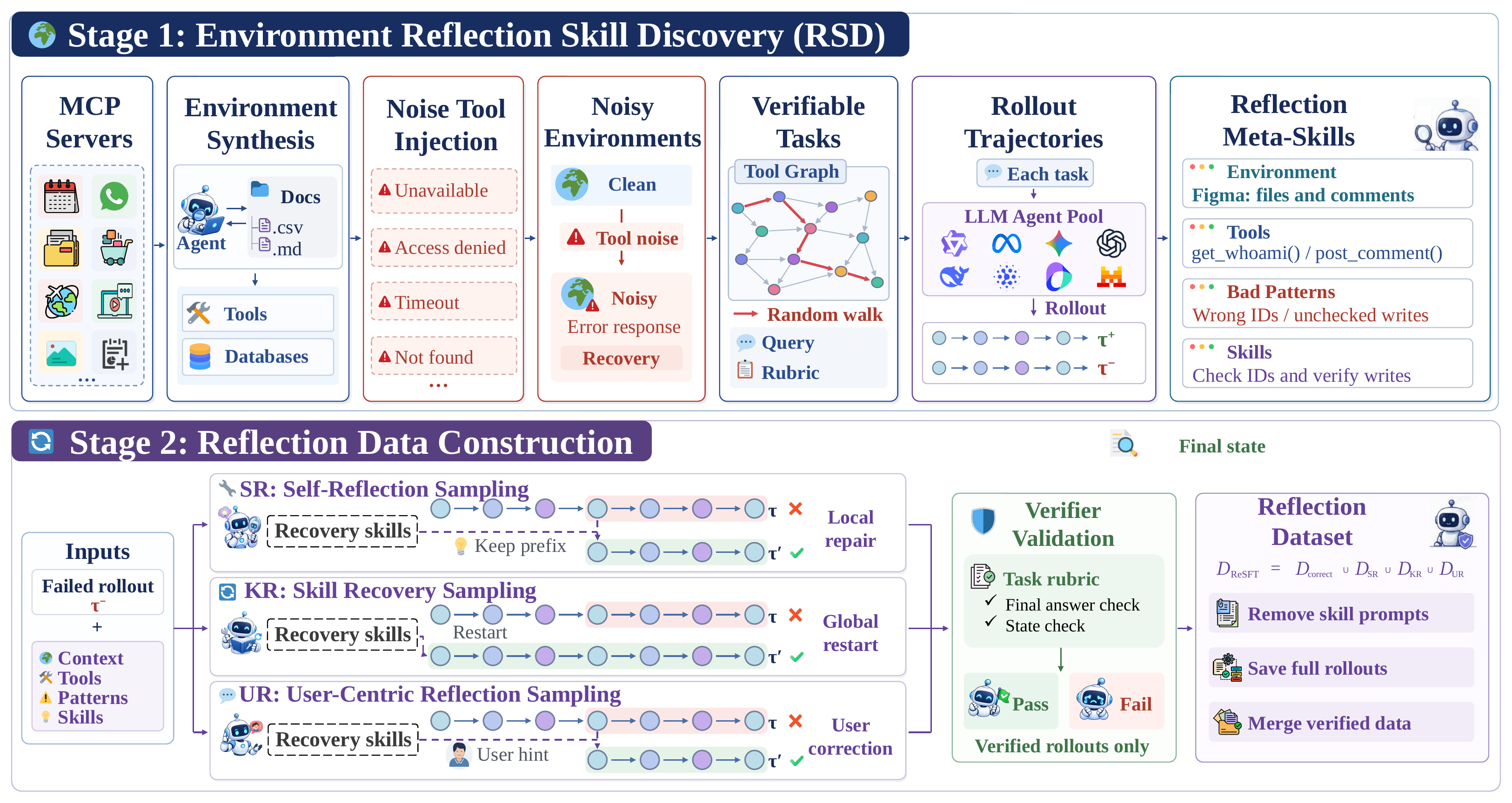}
  \caption{\textbf{\rsd{} and \resft{}.} \textbf{Stage 1} perturbs tool returns of 557 MCP-anchored executable environments with mined failure families, contrasting success/failure rollouts into an offline skill library. \textbf{Stage 2} samples local-repair (SR), global-recovery (KR), and user-correction (UR) continuations from that library, keeping only verifier-validated ones with skill prompts stripped.}
  \label{fig:overview}
\end{figure}

\textbf{Overview.} Figure~\ref{fig:overview} summarizes the pipeline. To convert
environment failures into a reflection capability, \method{} comprises three stages:

\textbf{(1)~Environment Reflection Skill Discovery (RSD) (\hyperref[sec:rsd]{§~\ref*{sec:rsd}})}. Build failure-aware executable environments and mine verifier-matched pairs into an offline skill library. 

\textbf{(2)~Reflective Supervised Fine-Tuning (ReSFT) (\hyperref[sec:resft]{§~\ref*{sec:resft}}).} Convert each skill into verifier-validated self-reflection, recovery, and user-correction data. 

\textbf{(3)~On-Policy Reflective Policy Optimization (ORPO) (\hyperref[sec:orpo]{§~\ref*{sec:orpo}}).} Combine Noisy Agent RL with same-token skill-guided OPD to provide dense reflection supervision. 

\subsection{Stage 1: Environment Reflection Skill Discovery (RSD)}
\label{sec:rsd}

\subsubsection{Noise-Grounded Environment Scaling}

 To construct realistic environments with tool failures, we select 557 high-usage MCP server specifications from Smithery\footnote{\url{https://smithery.ai/}}, each defining an environment topic \(m\in\mathcal{M}\). Following Agent-World~\citep{dong2026agentworld}, a research agent equipped with search, browser, coding, and operating-system tools builds and iteratively enriches a topic-grounded database. A coding agent then generates candidate tools and unit tests from each topic and its database. We retain only tools that compile, pass all tests, satisfy state-transition constraints, and support executable task verification. Thus, each environment \(\mathcal{E}_m\) consists of a database, a validated tool set, and corresponding verifiers. The resulting ecosystem is \(\mathcal{E}=\{\mathcal{E}_m:m\in\mathcal{M}\}\), with 60 verifier-backed tasks per topic; the resulting task set \(\mathcal{Q}\) supplies the ReSFT data.

\textbf{Tool-Failure Mining and Injection.}
Because real deployment is unstable, we ground perturbations in observed failures rather than invent them. We replay agents across nine public agentic benchmarks whose environments expose production-grade tool ecosystems and log every tool return (Appendix~\ref{app:mining-benchmarks}). In total we collect 177,700 tool returns, of which 24,510 are error returns, and group them by surface form into eight actionable failure families (Table~\ref{tab:noise-taxonomy}). We then inject these failures into tool returns probabilistically to emulate real deployment:
\begin{equation}
 o_{t+1}^{\mathrm{train}}=
 \begin{cases}
 J(o_{t+1}), & b_{t+1}^{\mathrm{inj}}=1,\\
 o_{t+1}, & b_{t+1}^{\mathrm{inj}}=0,
 \end{cases}
 \qquad b_{t+1}^{\mathrm{inj}}\sim\mathrm{Bernoulli}(p_{\mathrm{inj}}),
 \label{eq:training-injection}
\end{equation}
where the renderer $J$ replaces the true return with a failure drawn from those compatible with the
invoked tool $f_t$, balanced across the eight families so that rare ones still receive enough
instances for recovery to be learned. Here $p_{\mathrm{inj}}$ is the per-return injection probability, and we keep clean and perturbed
tasks only if they admit a verifiable solution and recovery, respectively.

\subsubsection{Reflection Meta-Skill Extraction}
\textbf{Capability-Diverse Trajectory Sampling and Pairing.}
Repeated sampling from one model can concentrate trajectories around its model-specific reasoning tendencies~\citep{jiang2023llmblender,du2024multiagentdebate}. We therefore sample trajectories from a capability-diverse LLM pool $\Pi_{\mathrm{pool}}$ of nine models spanning five families, ranging from 4B to 284B parameters (Table~\ref{tab:model-pool}). Each model interacts with the noisy ecosystem $\mathcal{E}$ on every task $q\in\mathcal{Q}$ under the same sampling protocol.

Because rule-based metrics cannot reliably assess long-horizon trajectories, an analyzer model and the environment verifier jointly label each rollout, yielding
$\mathcal{D}_{\mathrm{correct}}=\{(q_i,\tau_i^+)\}_{i=1}^{N_+}$
and
$\mathcal{D}_{\mathrm{error}}=\{(q_i,\tau_i^-,e_i,t_i^\star,s_{t_i^\star})\}_{i=1}^{N_-}$,
where $\tau_i^+$ and $\tau_i^-$ denote correct and erroneous trajectories, $e_i$ is the diagnosed error rationale, $t_i^\star$ is the first erroneous turn, and $s_{t_i^\star}$ is its environment-state snapshot. We pair correct and erroneous trajectories from the same task, limit the number of pairs contributed by each task, and require each environment to draw pairs from at least five distinct tasks, forming:
$\mathcal{D}_{\mathrm{pair}}=\{(q,\tau^+,\tau^-,e,t^\star,s_{t^\star})\}$.

\textbf{Meta-Skill Extraction.}
An ideal reflective meta-skill is an environment recovery rule that maps an agent's bad pattern to a corrective action, enabling reuse across tasks within the same environment. To mine such skills, an analyzer examines paired successful and failed trajectories, identifies recurring failure patterns, and converts them into executable recovery rules. For each environment $m$, we store
$\zeta_m=(d_m,\mathcal{T}_m,\mathcal{B}_m,\mathcal{K}_m)$,
where $d_m$ describes the environment and its business objects, $\mathcal{T}_m$ its tools, $\mathcal{B}_m$ its observed bad patterns, and $\mathcal{K}_m$ the extracted reflection skills. The resulting offline library is
$\mathcal{K}=\bigcup_{m\in\mathcal{M}}\mathcal{K}_m$,
so each sample in $\mathcal{D}_{\mathrm{pair}}$ is associated with a skill $\skill\in\mathcal{K}_m$. Representative environment-specific skills are shown in Appendix~\ref{app:skill-examples}.

\subsection{Stage 2: Reflective Supervised Fine-Tuning (ReSFT)}
\label{sec:resft}
ReSFT converts explicit reflection skills into policy behavior. Starting from a frozen behavior policy $\pi_0$, we construct three complementary forms of reflective supervision for local repair, global recovery, and user-guided correction.

\textbf{Self-Reflection Sampling (SR).}
For an erroneous trajectory $\tau^-$, let $t^\star$ be its first erroneous turn, $s_{t^\star}$ the environment-state snapshot taken before that turn is executed, $h_{t^\star}$ the deployment-visible history up to that turn, and $\skill$ the matched reflection skill. Let $x\oplus z$ denote appending $z$ to context $x$, and let $\operatorname{Ver}_{s}(x,y)$ indicate whether continuation $y$ successfully completes the task after restoring the environment to state $s$. To teach failure awareness and self-reflection, we restore the environment to $s_{t^\star}$ and continue after appending the corrective skill:
\begin{equation}
y^{\mathrm{SR}}\sim\pi_0(\cdot\mid h_{t^\star}\oplus\skill),\qquad
\mathcal{D}_{\mathrm{SR}}=
\{(h_{t^\star},y^{\mathrm{SR}}):
\operatorname{Ver}_{s_{t^\star}}(h_{t^\star},y^{\mathrm{SR}})=1\}.
\end{equation}
The skill triggers diagnosis and replanning during generation but is removed from the stored input, so the policy learns to recover from its own error state.

\textbf{Skill Recovery Sampling (KR).}
Many failures originate from an incorrect initial plan rather than a local tool error. To teach global recovery, we reset the environment to its initial state $s_0$ and restart from the original task with the matched skill:
\begin{equation}
y^{\mathrm{KR}}\sim\pi_0(\cdot\mid q\oplus\skill),\qquad
\mathcal{D}_{\mathrm{KR}}=
\{(q,y^{\mathrm{KR}}):
\operatorname{Ver}_{s_0}(q,y^{\mathrm{KR}})=1\}.
\end{equation}
As in SR, the privileged skill is removed before storage, encouraging recovery without an explicit skill prior.

\textbf{User-Centric Reflection Sampling (UR).}
When an agent deviates from the user's intent, users may intervene and correct its reasoning. To teach user-centric reflection, a user simulator $U$ converts the matched skill into a natural-language correction:
$u_{\skill}=U(\skill,h_{t^\star},s_{t^\star})$. We then restore the environment to $s_{t^\star}$, insert the correction after the erroneous turn, and continue:
\begin{equation}
y^{\mathrm{UR}}\sim\pi_0(\cdot\mid h_{t^\star}\oplus u_{\skill}),\qquad
\mathcal{D}_{\mathrm{UR}}=
\{(h_{t^\star}\oplus u_{\skill},y^{\mathrm{UR}}):
\operatorname{Ver}_{s_{t^\star}}(h_{t^\star}\oplus u_{\skill},y^{\mathrm{UR}})=1\}.
\end{equation}
Because the user message is deployment-visible, $u_{\skill}$ remains in the stored input, whereas the privileged skill $\skill$ is removed.

Let $\mathcal{D}_{\mathrm{correct}}^{\pi_0}\subseteq\mathcal{D}_{\mathrm{correct}}$ denote correct trajectories generated by the target policy $\pi_0$. We merge them with the three reflective datasets:
\begin{equation}
\mathcal{D}_{\mathrm{ReSFT}}=
\mathcal{D}_{\mathrm{correct}}^{\pi_0}\cup
\mathcal{D}_{\mathrm{SR}}\cup
\mathcal{D}_{\mathrm{KR}}\cup
\mathcal{D}_{\mathrm{UR}},
\qquad
\mathcal{L}_{\mathrm{ReSFT}}
=-\!\!\sum_{(x,y)\in\mathcal{D}_{\mathrm{ReSFT}}}
\log\pi_\theta(y\mid x).
\label{eq:resft}
\end{equation}

\subsection{Stage 3: On-Policy Reflective Policy Optimization (ORPO)}
\label{sec:orpo}
ReSFT provides an offline initialization by internalizing explicit reflection skills. To further elicit proactive reflection from the current policy on its own on-policy rollouts, ORPO supplements outcome-based agentic reinforcement learning with token-level, reflection-guided on-policy distillation (OPD)~\citep{agarwal2024gkd,zhao2026opsd,wu2026seed}, providing dense supervision for failure-causing decisions.

\begin{figure}[t]
  \centering
  \includegraphics[width=\linewidth]{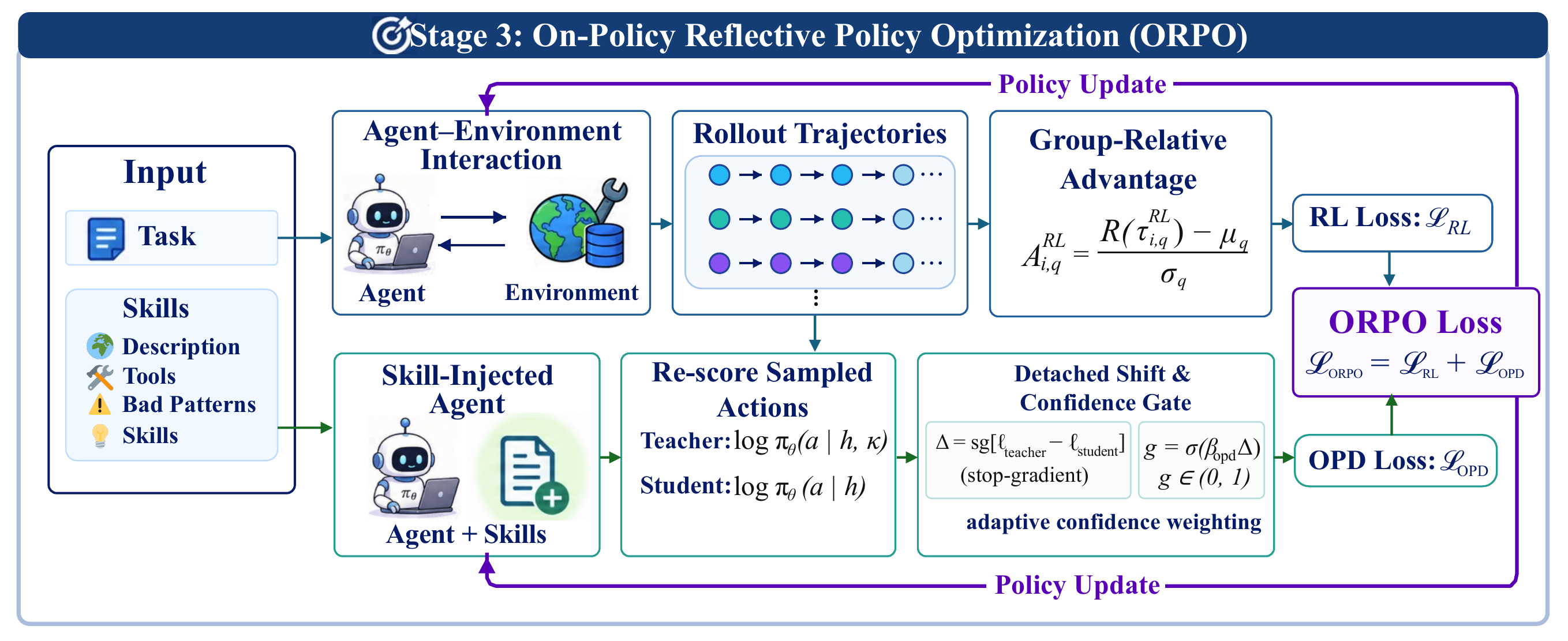}
\caption{\textbf{ORPO.} A mixed rollout group of mostly clean and a few perturbed trajectories shares one group-normalized advantage. Re-scoring the same tokens with and without the skill card of their environment yields the detached, confidence-weighted OPD signal.}
  \label{fig:afpo}
\end{figure}

\textbf{Noisy Agentic Reinforcement Learning.}
At update $u$, we draw a task $q$ from the EnvScaler RL task pool $\mathcal{Q}_{\mathrm{RL}}$~\citep{song2026envscaler}, which ORPO directly reuses instead of \rsd{}-synthesized tasks, and construct one mixed on-policy group to expose
the behavior policy to noisy environment feedback. Specifically, the policy
$\pi_{\theta_{\mathrm{old}}}$ generates $G$ trajectories
$\mathcal{G}_q=\{\tau_q^{(i)}\}_{i=1}^{G}$, with a fraction $\nu$ rolled out in noisy environments
and the remainder in clean ones. All trajectories retain the complete tool interaction and final
answer, and none is given a reflection-skill prompt. After rollout, the environment verifier assigns
each trajectory a return $R_{q,i}$; we normalize these returns within the mixed group:
\begin{equation}
\mu_q=\frac{1}{G}\sum_{i=1}^{G}R_{q,i},\qquad
\sigma_q^{2}=\frac{1}{G}\sum_{i=1}^{G}\bigl(R_{q,i}-\mu_q\bigr)^{2},\qquad
\widehat A_{q,i}=\frac{R_{q,i}-\mu_q}{\sigma_q+\epsilon},
\label{eq:adv-ratio}
\end{equation}
where normalizing over the mixed group rather than within its clean and perturbed strata keeps
genuine failures inside the on-policy distribution without letting them form a contrast of their own.
For an agent action token $a_{q,i,t,k}$ at position $k$ of turn $t$ under history $h_{q,i,t}$,
we define the token log-probability of the current policy and its ratio to the behavior policy:
\begin{equation}
\ell^{\theta}_{q,i,t,k}=\log\pi_\theta\bigl(a_{q,i,t,k}\mid h_{q,i,t},a_{q,i,t,<k}\bigr),\qquad
\rho_{q,i,t,k}(\theta)=\exp\bigl(\ell^{\theta}_{q,i,t,k}-\ell^{\mathrm{old}}_{q,i,t,k}\bigr),
\label{eq:logprobs}
\end{equation}
with $\ell^{\mathrm{old}}$ the same score under $\pi_{\theta_{\mathrm{old}}}$. Letting
$\chi_{q,i,t,k}\in\{0,1\}$ mask the valid agent-generated tokens, which excludes tool returns and
simulated user messages, the clipped Agent RL objective is
\begin{multline}
\mathcal{L}_{\mathrm{RL}}(\theta)=-\,\mathbb{E}_{q,i,t,k}\Bigl[\chi_{q,i,t,k}\min\bigl(\rho_{q,i,t,k}\widehat A_{q,i},\
\operatorname{clip}(\rho_{q,i,t,k},1-\epsilon_c,1+\epsilon_c)\,\widehat A_{q,i}\bigr)\Bigr]\\
+\beta_{\mathrm{KL}}\,\mathbb{E}_{h\sim\mathcal{G}_q}\bigl[D_{\mathrm{KL}}\bigl(\pi_\theta(\cdot\mid h)\Vert\pi_{\mathrm{ref}}(\cdot\mid h)\bigr)\bigr],
\label{eq:rl-hybrid}
\end{multline}
with $\epsilon,\epsilon_c,\beta_{\mathrm{KL}}$ hyperparameters and $\mathbb{E}_{q,i,t,k}$ the mean over
all sampled token positions of the group. This component propagates only the outcome supervision of
the verifier; the reflection-specific token-level signal comes next.

\textbf{Reflection On-Policy Distillation.}
To turn the sparse outcome signal into dense supervision over reflective decisions, we condition on
the reflection meta-skills $\skill_{q,i}\in\mathcal{K}_m$ of the environment $m$ that task $q$ instantiates.
No retrieval or extra matching step is involved: each EnvScaler RL task already specifies its environment, and the skill library is organized by environment, so the
corresponding skill card is loaded directly with the environment and never depends on a benchmark answer. Prompting the same policy $\pi_\theta$ with
this skill prepended to the task turns it into a teacher that re-scores the student rollout sampled
without any skill prior: a deterministic function $H$ builds the skill-augmented history
$\widetilde h_{q,i,t}=H(h_{q,i,t},\skill_{q,i})$, and the current policy re-scores the already sampled
action token under it,
\begin{equation}
\ell^{\mathrm{skill}}_{q,i,t,k}=\log\pi_\theta\bigl(a_{q,i,t,k}\mid\widetilde h_{q,i,t},a_{q,i,t,<k}\bigr),
\label{eq:opd-logprob}
\end{equation}
which is contrasted with the ordinary-context score $\ell^{\theta}_{q,i,t,k}$ of
Equation~\ref{eq:logprobs} on exactly the same token. OPD therefore needs a single environment
rollout stream: the skill branch only scores tokens as a training-time privileged teacher, and
gradients flow through the ordinary branch alone. Following SEED, the detached skill-induced shift
and its confidence weight are
\begin{equation}
\Delta_{q,i,t,k}=\operatorname{sg}\bigl[\ell^{\mathrm{skill}}_{q,i,t,k}-\ell^{\theta}_{q,i,t,k}\bigr],\qquad
\gamma_{q,i,t,k}=\operatorname{sigmoid}\bigl(\beta_{\mathrm{opd}}\Delta_{q,i,t,k}\bigr),
\end{equation}
and we optimize
\begin{equation}
\mathcal{L}_{\mathrm{OPD}}(\theta)=\mathbb{E}_{q,i,t,k}\Bigl[\chi_{q,i,t,k}\,\gamma_{q,i,t,k}\bigl(\operatorname{sg}[\ell^{\mathrm{skill}}_{q,i,t,k}]-\ell^{\theta}_{q,i,t,k}\bigr)\Bigr],
\label{eq:opd}
\end{equation}
where $\operatorname{sg}[\cdot]$ blocks gradients through the teacher branch, so minimizing
$\mathcal{L}_{\mathrm{OPD}}$ raises the student score on teacher-endorsed on-policy tokens. The
confidence weight $\gamma\in(0,1)$ is a soft weighting rather than a hard filter: it emphasizes tokens
with a positive skill-induced shift and down-weights tokens whose shift is weak or negative.

\begin{table*}[!t]
    \centering
    \caption{\textbf{Main results on five tool-use benchmarks.} Overall is the unweighted mean of the five benchmark averages; Avg.\ of BFCL V4 and ACEBench follows each benchmark's official scoring rather than a simple mean. Best and second-best among 7B--8B methods are \textbf{bolded} and \underline{underlined} (ties share a mark). $^*$Reproduced by us because the checkpoint is not open-sourced.}
    \label{tab:main-results}

    \vspace{0.35em}
    \setlength{\tabcolsep}{1.0pt}
    \fontsize{6.1pt}{8.2pt}\selectfont
    \renewcommand{\arraystretch}{1.18}

    \resizebox{\textwidth}{!}{%
    \begin{tabular}{@{}l*{5}{c}*{4}{c}*{7}{c}*{3}{c}cc@{}}
    \toprule

    \multirow{2}{*}{\textbf{Method}}
     & \multicolumn{5}{c}{\textbf{VitaBench}}
     & \multicolumn{4}{c}{\textbf{$\tau^2$-Bench}}
     & \multicolumn{7}{c}{\textbf{BFCL V4}}
     & \multicolumn{3}{c}{\textbf{ACEBench}}
     & \multirow{2}{*}{\textbf{\shortstack{Tool\\Sandbox}}}
     & \multirow{2}{*}{\textbf{Overall}} \\

    \cmidrule(lr){2-6}
    \cmidrule(lr){7-10}
    \cmidrule(lr){11-17}
    \cmidrule(lr){18-20}

     & Cro. & Del. & In. & OTA. & Avg.
     & Ret. & Air. & Tel. & Avg.
     & Web. & Mem. & Mul. & NoL. & Liv. & Irr. & Avg.
     & Ste. & Tur. & Avg. & & \\

    \midrule

    \multicolumn{22}{l}{\textbf{\textit{Foundation Models}}}\\
    \addlinespace[1pt]

    GPT-5.2
     & 24.3 & 52.5 & 45.8 & 28.3 & 37.7
     & 81.6 & 62.5 & 95.8 & 80.0
     & 75.5 & 45.8 & 48.5 & 81.9 & 70.4 & 88.7 & 62.9
     & 95.0 & 50.0 & 68.0 & 73.3 & 64.4 \\

    Claude-4.5-Sonnet
     & 23.5 & 52.0 & 48.3 & 28.3 & 38.0
     & 86.2 & 70.1 & 98.0 & 84.7
     & 81.0 & 65.0 & 61.4 & 88.7 & 81.1 & 86.6 & 73.2
     & 100.0 & 70.0 & 82.0 & 85.0 & 72.6 \\

    Gemini 3 Pro
     & 31.5 & 62.0 & 56.5 & 40.3 & 47.6
     & 85.3 & 72.7 & 98.0 & 85.4
     & 80.0 & 61.7 & 60.8 & 90.7 & 83.1 & 85.6 & 72.5
     & 100.0 & 50.0 & 70.0 & 88.4 & 72.8 \\

    Qwen3-235B-A22B
     & 14.5 & 45.0 & 32.0 & 15.8 & 26.8
     & 71.9 & 45.6 & 58.0 & 58.5
     & 54.0 & 23.9 & 45.4 & 37.4 & 68.9 & 81.7 & 48.0
     & 100.0 & 63.3 & 78.0 & 77.1 & 57.7 \\

    Qwen3-8B
     & 1.2 & 15.5 & 24.8 & 4.0 & 11.4
     & 34.0 & 26.5 & 18.0 & 26.2
     & 7.0 & 17.6 & 35.4 & 90.2 & 80.9 & 77.2 & 40.4
     & 80.0 & 36.7 & 54.0 & 69.1 & 40.2 \\

    \midrule

    \multicolumn{22}{l}{\textbf{\textit{Reflection Methods}} (8B)}\\
    \addlinespace[1pt]

    Reflexion-8B
     & 1.0 & 13.0 & 11.0 & 2.0 & 6.8
     & 43.0 & 36.0 & 24.6 & 34.5
     & 3.5 & 20.9 & \underline{46.5} & 86.8 & 80.2 & 72.1 & 42.7
     & \underline{75.0} & 56.7 & 64.0 & 75.4 & 44.7 \\

    ExpeL-8B
     & 0.0 & 14.0 & 11.0 & 4.0 & 7.3
     & 45.6 & 34.0 & 21.1 & 33.6
     & 2.5 & 23.0 & 39.5 & \underline{88.8} & 78.2 & 80.2 & 41.7
     & 55.0 & 36.7 & 44.0 & 63.0 & 37.9 \\

    Agent-R-8B$^*$
     & 2.0 & 15.0 & 8.0 & 4.0 & 7.3
     & 37.7 & 32.0 & 21.9 & 30.6
     & 1.0 & 17.9 & 35.4 & 87.6 & 80.2 & 74.7 & 38.7
     & 60.0 & 43.3 & 50.0 & 73.7 & 40.1 \\

    NoisyAgent-8B$^*$
     & \underline{5.0} & 22.0 & 27.0 & 7.0 & 15.3
     & 47.6 & 40.0 & 40.8 & 42.8
     & 2.5 & 13.3 & 34.8 & 65.4 & 79.9 & 74.6 & 35.6
     & 50.0 & 53.3 & 52.0 & 71.5 & 43.4 \\

    \midrule

    \multicolumn{22}{l}{\textbf{\textit{Environment-Scaling Methods}} (7B--8B)}\\
    \addlinespace[1pt]

    Simulator-8B
     & 0.0 & 1.2 & 4.8 & 1.0 & 1.8
     & 32.2 & 34.0 & 29.2 & 31.8
     & 17.5 & 6.0 & 4.1 & 47.6 & 44.6 & \textbf{87.3} & 23.9
     & 35.0 & 30.0 & 32.0 & 61.7 & 30.2 \\

    TOUCAN-7B
     & 0.5 & 6.9 & 3.0 & 0.8 & 2.8
     & 22.8 & 20.0 & 10.5 & 17.7
     & 21.0 & 18.5 & 17.8 & 81.0 & 73.9 & 78.6 & 36.6
     & 70.0 & 26.7 & 44.0 & 61.9 & 32.6 \\

    AWM-8B
     & 1.5 & 16.0 & 19.2 & 4.2 & 10.2
     & 41.2 & 23.5 & 38.5 & 34.4
     & 9.5 & 15.7 & 34.9 & \textbf{90.2} & 80.5 & 73.9 & 40.0
     & 60.0 & 46.7 & 52.0 & 71.5 & 41.6 \\

    EnvACE-8B$^*$
     & \textbf{6.0} & \underline{24.0} & 27.0 & 7.0 & \underline{16.0}
     & 48.9 & \underline{44.0} & 17.3 & 36.7
     & 12.3 & \textbf{24.0} & 45.3 & 87.6 & 81.2 & \underline{83.2} & 46.0
     & 55.0 & 43.3 & 48.0 & 71.6 & 43.7 \\

    EnvScaler-8B
     & \textbf{6.0} & 20.8 & \textbf{29.5} & 7.0 & 15.8
     & 49.6 & 31.5 & 32.7 & 37.9
     & 23.0 & 21.9 & \textbf{47.1} & 88.5 & \textbf{82.2} & 74.6 & 47.6
     & \textbf{85.0} & 60.0 & \underline{72.5} & \underline{80.5} & 50.9 \\

    Agent-World-8B$^*$
     & \textbf{6.0} & 21.0 & 11.0 & \textbf{20.0} & 14.5
     & \textbf{72.8} & 40.0 & \underline{50.9} & \underline{54.6}
     & \textbf{47.0} & 21.7 & 44.5 & 83.3 & 79.6 & 80.2 & \underline{51.4}
     & \textbf{85.0} & \underline{63.3} & 72.0 & 75.6 & \underline{53.6} \\

    \rowcolor[HTML]{F0E6FF}
    \textbf{Agent-Reflex-8B (ours)}
     & \textbf{6.0} & \textbf{27.0} & \underline{29.0} & \underline{8.0} & \textbf{17.5}
     & \underline{68.4} & \textbf{46.0} & \textbf{71.9} & \textbf{62.1}
     & \underline{44.5} & \underline{23.5} & \textbf{47.1} & \underline{88.8} & \underline{81.4} & 81.4 & \textbf{52.9}
     & \textbf{85.0} & \textbf{73.3} & \textbf{78.0} & \textbf{82.5} & \textbf{58.6} \\

    \bottomrule
    \end{tabular}%
    }
\end{table*}

\textbf{Joint Objective.}
We jointly optimize the outcome-based Agent RL and reflection OPD objectives:
\begin{equation}
\mathcal{L}_{\mathrm{ORPO}}=\mathcal{L}_{\mathrm{RL}}+\lambda_{\mathrm{opd}}\mathcal{L}_{\mathrm{OPD}}.
\label{eq:orpo}
\end{equation}
The objective combines trajectory-level outcome learning with token-level reflection supervision,
further sharpening the proactive reflection of the policy.

\section{Experiments}
\label{sec:experiments}
\subsection{Experimental Setups}
\textbf{Datasets.} We evaluate our method on two groups of benchmarks. (1)~\textbf{Long-horizon tool use} covers general function calling and interactive workflows: BFCL-v4~\citep{patil2025bfcl}, ACEBench~\citep{chen2025acebench}, 200 stratified ToolSandbox tasks~\citep{lu2025toolsandbox}, $\tau^2$-Bench~\citep{barres2025tau2}, and VitaBench~\citep{he2025vitabench}. (2)~\textbf{Noisy interaction} uses AgentNoiseBench~\citep{wang2026agentnoisebench}, instantiated on the $\tau^2$-Bench and VitaBench tool- and user-noise settings. Following the EnvScaler protocol~\citep{song2026envscaler}, all evaluations use the standard MCP format and task templates, and each benchmark is scored with its official metric. Appendix~\ref{app:datasets} specifies every benchmark.

\textbf{Baselines.} Every trained method here, including \method{}, starts from the same Qwen3-8B backbone under one fair experimental setting. We compare three families: \textit{reflection methods}~\citep{shinn2023reflexion,zhao2024expel,yuan2025agentr,chen2026noisyagent}, \textit{environment-scaling methods}~\citep{li2025simia,xu2025toucan,song2026envscaler,wang2026awm,xu2026envace,dong2026agentworld}, and \textit{trajectory-level RL algorithms}~\citep{shao2024deepseekmath,yu2025dapo,hu2025reinforce++,wu2026seed}. Appendix~\ref{app:baselines} specifies every baseline and Appendix~\ref{app:implementation} the implementation.

\subsection{Main Results}

\subsubsection{Results on Long-Horizon Tool Use.}
Table~\ref{tab:main-results} shows that \method{} performs well on five long-horizon agent benchmarks, achieving the highest overall score among strong baselines. We further make the following observations.

\textbf{(1) Internalizing Reflection beyond Inference-Time Guidance.}
\method{} outperforms Reflexion-8B and ExpeL-8B on VitaBench, ACEBench, and ToolSandbox. On VitaBench, \method{} achieves 17.5\%, compared with 6.8\% for Reflexion-8B. While these methods rely on self-generated critiques or retrieved experience at inference time, \method{} trains directly on verified recovery behavior. This may help the policy translate reflection into corrective actions more reliably, reducing its reliance on interpreting textual advice during execution.

\textbf{(2) Reusable Recovery Knowledge beyond Individual Trajectories.}
\method{} surpasses Agent-R-8B and NoisyAgent-8B on VitaBench, BFCL, ACEBench, and ToolSandbox. On ACEBench, it reaches 78.0\%, compared with 52.0\% for NoisyAgent-8B. These gains suggest that environment-level meta-skills provide useful recovery guidance beyond individual correction trajectories or episode-level rewards.

\textbf{(3) Reflective Supervision Complements Environment Scaling.}
\method{} improves over EnvScaler-8B and Agent-World-8B on all five benchmark averages, e.g., 62.1\% vs.\ 54.6\% on $\tau^2$-Bench. On ACEBench, the gain comes mainly from multi-turn interaction: \method{} reaches 73.3\% at the turn level while matching the baselines at 85.0\% step-level accuracy. This suggests that reflective supervision complements broader tool and task coverage.

\begin{table*}[t]
\centering
\caption{\textbf{Results on AgentNoiseBench.} We report performance under tool and user noise. Avg.\ is the mean of the tool-noise and user-noise scores, and Overall is the mean of the two benchmark averages. The \textit{Clean} results of foundation models are taken from the original AgentNoiseBench report, so they may differ slightly from Table~\ref{tab:main-results}. $^*$ marks our reproduction.}
\label{tab:noise-results}

\resizebox{\textwidth}{!}{%
\begin{tabular}{@{}l>{\color{black!55}}c c c c
>{\color{black!55}}c c c c c@{}}
\toprule

\multirow{2}{*}{\textbf{Method}}
& \multicolumn{4}{c}{\textbf{AgentNoiseBench-$\tau^2$}}
& \multicolumn{4}{c}{\textbf{AgentNoiseBench-Vita}}
& \multirow{2}{*}{\textbf{Overall}} \\

\cmidrule(lr){2-5}
\cmidrule(lr){6-9}

& \textit{Clean} & Tool Noise & User Noise & Avg.
& \textit{Clean} & Tool Noise & User Noise & Avg.
& \\

\midrule

\multicolumn{10}{l}{\textit{Foundation Models}}\\
\addlinespace[1pt]

GPT-5.2
& 81.0 & \textbf{61.0} & \textbf{72.0} & \textbf{66.5}
& 53.0 & \textbf{45.0} & \textbf{45.0} & \textbf{45.0}
& \textbf{55.8} \\

Claude-4.5-Sonnet
& 73.0 & \underline{57.0} & \underline{65.0} & \underline{61.0}
& 51.0 & \underline{41.0} & \underline{41.0} & \underline{41.0}
& \underline{51.0} \\

Gemini-2.5-Pro
& 60.0 & 37.0 & 55.0 & 46.0
& 49.0 & 34.0 & 38.0 & 36.0
& 41.0 \\

Qwen3-8B
& 26.2 & 14.1 & 26.2 & 20.2
& 11.4 & 3.3 & 5.5 & 4.4
& 12.3 \\

\midrule

\multicolumn{10}{l}{\textit{Reflection and Environment Scaling (Qwen3-8B)}}\\
\addlinespace[1pt]

Reflexion-8B
& 34.5 & 17.2 & 29.3 & 23.3
& 6.8 & \underline{7.8} & 0.2 & 4.0
& 13.7 \\

Agent-R-8B$^{\ast}$
& 30.6 & 14.5 & 25.4 & 20.0
& 7.3 & 3.9 & 0.8 & 2.4
& 11.2 \\

NoisyAgent-8B
& 42.8 & -- & -- & \underline{37.6}
& 15.3 & -- & -- & \textbf{14.2}
& \underline{25.9} \\

AWM-8B
& 34.4 & 14.1 & 22.6 & 18.4
& 10.2 & 3.2 & 0.6 & 1.9
& 10.2 \\

EnvScaler-8B
& 37.9 & \underline{20.8} & \underline{33.4} & 27.1
& 15.8 & 5.0 & \underline{13.6} & 9.3
& 18.2 \\

\rowcolor[HTML]{F0E6FF}
\textbf{Agent-Reflex-8B (ours)}
& 62.1 & \textbf{38.4} & \textbf{53.4} & \textbf{45.9}
& 17.5 & \textbf{9.4} & \textbf{15.2} & \underline{12.3}
& \textbf{29.1} \\

\bottomrule
\end{tabular}%
}
\end{table*}

\subsubsection{Robustness under Interaction Noise.}
To assess whether \method{} remains effective when tool feedback or user interactions are perturbed, we evaluate it on AgentNoiseBench. In detail, every foundation model degrades markedly.

\textbf{Generalization of \method{} to Unseen Perturbations.}
Table~\ref{tab:noise-results} shows that \method{} achieves the highest Overall score among the compared 8B methods, reaching 29.1\% compared with 25.9\% for NoisyAgent-8B and outperforming the other strong baselines overall. These results indicate that its effectiveness extends to noisy interaction. Appendix~\ref{app:reflective} provides a fine-grained attribution of the observed noise-robustness gains.

\begin{figure}[t]
\centering
\includegraphics[width=\textwidth]{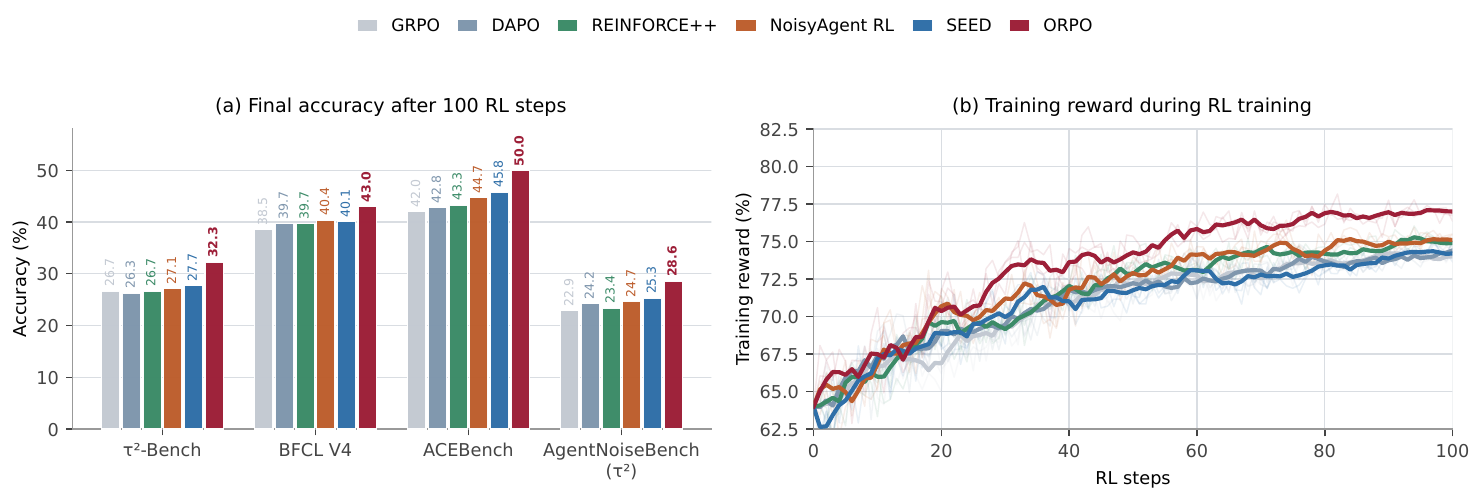}
\caption{\textbf{Comparison of agent policy optimization algorithms.} All runs start from Qwen3-8B and train for 100 RL steps on the same task pool under matched budgets. \textbf{(a)} Final accuracy on held-out benchmarks. \textbf{(b)} Mean training reward on the same tasks.}
\label{fig:algo-compare}
\end{figure}

\subsection{Does On-Policy Reflective Optimization Improve Robustness?}
To further assess ORPO's effectiveness for reflection training, we analyze along two dimensions.

\textbf{(1) Performance Comparison.} To evaluate online reflective learning, we train five representative algorithms and ORPO from Qwen3-8B for 100 RL steps on the same task pool. In Figure~\ref{fig:algo-compare}(a), ORPO holds a stable advantage over GRPO, DAPO, and REINFORCE++ on BFCL V4, $\tau^2$-Bench, and real interaction noise. The advantage also holds against SEED, which already carries an OPD signal, and against noisy Agent RL, so reflection meta-skills summarized from real failures transfer better than experience estimated from rollouts alone.

\textbf{(2) Training Dynamics.} To assess ORPO's stability, Figure~\ref{fig:algo-compare}(b) tracks mean rollout reward over the same 100 steps. ORPO separates from GRPO and SEED early and keeps rising after both flatten. We attribute this stability to the confidence weighting, which emphasizes tokens with a positive skill-induced shift and down-weights those with weak or negative shifts, so skill-neutral tokens do not dominate the update.

\subsection{Ablation Study}
\begin{wraptable}{r}{0.44\textwidth}
\vspace{-\intextsep}
\centering
\vspace{-0.5em}
\captionsetup{skip=3pt}
\caption{\textbf{Component ablation.} AgentNoiseBench averages per variant; shading is the within-column normalized drop, so darker marks a more load-bearing module.}
\label{tab:ablation}
\setlength{\tabcolsep}{11pt}
\fontsize{7.8pt}{9.2pt}\selectfont
\renewcommand{\arraystretch}{1.26}
\begin{tabular}{l cc}
\toprule
\textbf{Variant} & \textbf{ANB-$\tau^2$} & \textbf{ANB-Vita} \\
\midrule
\textbf{\method{}} & \cellcolor{oursrow}\textbf{45.9} & \cellcolor{oursrow}\textbf{12.3} \\
\addlinespace[1.5pt]
w/o SR   & \cellcolor{skillpurple!75}\textcolor{white}{41.3} & \cellcolor{skillpurple!75}\textcolor{white}{8.9} \\
w/o KR   & \cellcolor{skillpurple!60}42.4 & \cellcolor{skillpurple!51}10.2 \\
w/o UR   & \cellcolor{skillpurple!52}43.0 & \cellcolor{skillpurple!38}10.9 \\
w/o OPD  & \cellcolor{skillpurple!42}43.7 & \cellcolor{skillpurple!64}\textcolor{white}{9.5} \\
w/o weight & \cellcolor{skillpurple!31}44.5 & \cellcolor{skillpurple!47}10.4 \\
\bottomrule
\end{tabular}
\vspace{-1em}
\end{wraptable}

Table~\ref{tab:ablation} reports an ablation study of
\method{} on the two AgentNoiseBench settings. Each row removes
one component while keeping the remaining training and evaluation
conditions fixed, and the complete-model scores match those in
Table~\ref{tab:noise-results}. We make three observations.
\textbf{(1)} Removing any component degrades both settings, showing
that the components are complementary. 
\textbf{(2)} The supervision types follow a consistent ordering:
removing SR causes the largest drop, followed by KR and UR,
suggesting that failure awareness and local repair form the
foundation of reflective behavior.
\textbf{(3)} The skill library alone is insufficient: with the
library retained, removing OPD clearly hurts performance, while
removing only its confidence weighting also causes a decline.
Thus, the gains arise from the token-level reflection signal rather
than skill exposure alone.

\section{Conclusion}
In this paper, we present \method{}, a general training framework for incentivizing proactive reflective thinking in agents. In detail, \rsd{} mines reusable reflection meta-skills from matched success--failure trajectories under grounded tool-feedback perturbations. Building on this skill library, \resft{} turns each skill into complementary supervision for self-reflection, skill recovery, and user-centric correction, and ORPO further elicits proactive reflection by combining agentic reinforcement learning under noisy interaction with reflection-guided on-policy distillation. Across six challenging agent benchmark suites, \method{} improves long-horizon tool use and robustness under tool and user noise with stable training dynamics, supporting scalable agent training.

\section*{AI use statement}

We used generative AI tools solely to assist with literature reading and to proofread and polish author-written text. These tools were not used to develop research ideas, design experiments, conduct analyses, generate results or citations, formulate survey or interview questions, or create or modify figures. We reviewed all AI-assisted content and take full responsibility for the final manuscript, including its claims and artifacts.

\section*{Reproducibility statement}

Readers can find information supporting the reproducibility of this work in
Appendix~\ref{app:implementation}. It includes detailed information on environment construction, failure mining and noise injection, trajectory sampling and skill extraction, ReSFT and ORPO training procedures, model and optimization settings, evaluation protocols, and contamination controls.

\FloatBarrier

\bibliographystyle{plainnat}
\bibliography{references}

\appendix
\newpage
\section*{Appendix}
\label{app:contents}
\startcontents[sections]
\setupappendixtoc
{\hypersetup{linkcolor=AppendixTOCSection}
\printcontents[sections]{l}{1}{\setcounter{tocdepth}{2}}}
\newpage

\section{Supplement Related Work}
\label{app:related-work}

\subsection{Agent Reflection Learning}
\label{app:rw-reflection}
Agent reflection studies how interaction feedback can reveal errors and guide subsequent behavior. The methods differ along one axis that matters for our claims: \emph{where the corrective signal lives}: in the context at inference time, in an offline trajectory rewrite, or in the parameters.

\textbf{Inference-time reflection.} The first family leaves parameters untouched and reflects in context. Reflexion appends a verbal self-critique to an episodic buffer that later attempts read~\citep{shinn2023reflexion}, and Self-Refine iterates generation, self-feedback, and revision within a single episode~\citep{madaan2023selfrefine}; ExpeL instead accumulates cross-task natural-language insights and retrieves similar past trajectories at test time~\citep{zhao2024expel}. These methods bound what prompting alone recovers, and that bound is informative: because the critique is produced by the same policy that failed, it is only as good as that policy's model of the environment. When a tool return is well-formed but wrong, there is nothing in the context to contradict the failed plan, so the critique tends to restate it. Our Table~\ref{tab:main-results} shows this concretely: both Reflexion and ExpeL fall below their own backbone on the longest-horizon benchmark.

\textbf{Learning from correction trajectories.} The second family moves correction into training. Agent-R combines MCTS with model-guided selection of the revision point to construct recovery trajectories, then learns timely correction by iterative self-training~\citep{yuan2025agentr}. ReflectRL transfers reflection elicited on golden negative trajectories into direct reasoning~\citep{bi2026reflectrl}, and LoongReflect combines privileged reflection supervision with outcome-based RL~\citep{zhang2026loongreflect}, the closest prior work to our ORPO stage in supervision \emph{type}, though its privileged context is an episode-level hindsight rewrite rather than an environment-level artifact. The shared limitation is that the correction is synthesized per trajectory: each repair is fitted to the episode that produced it, so nothing accumulates into a statement about the environment that a later, differently-failing episode can reuse.

\textbf{Robustness to perturbed interaction.} A third line attacks the problem from the environment side. NoisyAgent perturbs user instructions and tool responses, mixes clean and noisy rollouts, and adaptively raises noise difficulty, giving a complementary route to recovery and ambiguity handling~\citep{chen2026noisyagent}. This exposes the policy to failure states, which trajectory-level RL alone does not, but the supervision remains a scalar return: it can express that a perturbed episode was harder, not which recovery the specific failure admitted.

\textbf{Positioning.} \method{} shares the goal of learning from failure and differs in what it learns \emph{from} and what it stores. Before any policy training, it consolidates cross-model behavioral contrasts (matched success/failure pairs on the same task under the same injected noise) into explicit, environment-indexed reflection meta-skills. A skill is therefore reusable across episodes and mined once per environment, rather than rewritten per trajectory. These skills then supply supervision for autonomous recovery and user-directed correction (\resft{}) and a privileged teacher context for token-level distillation (ORPO). Because the skill is privileged only during training and stripped from every stored input, inference needs no skill prompt and no auxiliary reflector, unlike the inference-time family above, where the critique must be produced online at deployment.

\subsection{Scalable Environment Synthesis for Agent Training}
Existing approaches split into two families. The first uses language models as implicit textual world models: Simia generates environment feedback conditioned on tool specifications and interaction histories, and GenEnv couples generative simulation with a capability-adapted curriculum~\citep{li2025simia,guo2025genenv}. This removes most environment-engineering cost, but the fidelity of a simulated observation is bounded by the simulator's ability to keep state consistent, so long rollouts can drift away from any realizable environment. The second family makes dynamics explicit through executable tools, databases, and programmatic verifiers. EnvScaler, AWM, and ScaleEnv construct environments whose transitions execute and whose outcomes are checkable~\citep{song2026envscaler,wang2026awm,tu2026scaleenv}; Agent-World mines databases from real service themes, validates generated tools against them, and lets environments and policies co-evolve through continuous task generation~\citep{dong2026agentworld}; EnvACE instead internalizes dynamics by \emph{world rehearsal}, with one policy emitting both tool actions and the responses they induce~\citep{xu2026envace}; and TOUCAN takes a data-centric route, synthesizing tool-use trajectories from real MCP services~\citep{xu2025toucan}.

These efforts are why a large pool of executable environments can be assumed today. What they do not address is the status of \emph{failure} inside those environments: a verifier returning zero says an episode was hard, not which of dozens of tool calls broke the plan, what evidence exposed it, or which recovery the situation admitted. \method{} adopts the database-backed executable paradigm from this line and targets that residual.

\subsection{Agentic Reinforcement Learning}
GRPO removes the value network by normalizing rewards within a sampled group~\citep{shao2024deepseekmath}, and DAPO and REINFORCE++ improve stability and sample efficiency through clip asymmetry, dynamic sampling, token-level loss aggregation, and variance reduction~\citep{yu2025dapo,hu2025reinforce++}. Applying these to tool-using agents introduces a difficulty single-turn reasoning lacks: a long episode's return is one scalar attached to a decision sequence of unknown length. One line of work therefore reshapes the rollout or the credit assignment rather than the optimizer, through hierarchical rewards with a self-critic stage~\citep{dong2025toolstar}, entropy-triggered partial rollouts with soft advantage attribution~\citep{dong2025arpo}, and balanced exploration across the agent and environment branches~\citep{dong2025aepo}. A complementary route leaves the reward untouched and densifies the gradient instead: on-policy distillation scores the student's own sampled tokens under a teacher context, which SEED instantiates for agents by re-scoring the same action token with and without privileged information and gating a stop-gradient log-likelihood shift through a sigmoid~\citep{wu2026seed}. ORPO is orthogonal to the first route and reuses SEED's estimator unchanged in the second; what it contributes is the content of the privileged slot, an environment-level skill card mined once from empirical tool failures rather than a larger model or a per-episode hindsight rewrite.

\section{Datasets}
\label{app:datasets}
All evaluations use the standard MCP interaction format and each benchmark's official metric, reported by domain.

\subsection{General Function Calling}
\begin{itemize}[leftmargin=1em,itemsep=2pt]
\item \textbf{BFCL-v4}~\citep{patil2025bfcl} extends the Berkeley Function Calling Leaderboard from single-call accuracy toward agentic evaluation. We report Web, Memory, Multi-Turn, Non-Live, Live, and Irrelevance, and the average follows the official BFCL scoring procedure rather than a simple mean of the six categories. It is useful here because it separates call formatting from execution: a model can be near-saturated on Non-Live while failing the multi-turn categories.
\item \textbf{ACEBench}~\citep{chen2025acebench} evaluates normal, special, and agentic tool use, emphasizing when to call a tool and how to handle underspecified requests. We report step-level and turn-level accuracy, and the average follows the official ACEBench scoring procedure rather than a simple mean of the two; the turn-level metric credits a turn only when every call in it is correct, making it sensitive to error propagation.
\item \textbf{ToolSandbox}~\citep{lu2025toolsandbox} is stateful and conversational, with implicit tool state dependencies, an LLM user simulator, and deliberately unsolvable tasks that require asking rather than acting. We use a stratified 200-task subset and release the task identifiers so the subset is reproducible.
\end{itemize}

\subsection{Long-Horizon Interactive Workflows}
\begin{itemize}[leftmargin=1em,itemsep=2pt]
\item \textbf{$\tau^2$-Bench}~\citep{barres2025tau2} is \emph{dual-control}: agent and simulated user both act on shared state. We report Retail, Airline, and Telecom and their macro-average. Each domain supplies a policy document the agent must respect, and success requires the final database state to match the target, so partially correct trajectories score zero.
\item \textbf{VitaBench}~\citep{he2025vitabench} covers delivery, in-store, and OTA domains plus composite \emph{cross}-scenario tasks spanning domains within one episode. We report the four domains under official Avg@4 and take the benchmark average as the macro-average of the per-domain scores.
\end{itemize}

\subsection{Noisy Interaction}
\textbf{AgentNoiseBench}~\citep{wang2026agentnoisebench} injects perturbations into both sides of the interaction loop on the $\tau^2$-Bench and VitaBench environments. Tool noise covers faulty, incomplete, inducing, redundant, and wrong returns; user noise covers ambiguity, topic shift, conflicting instructions, redundancy, and boundary probing. We score with the official Stability-Gated Success metric, which credits a task only when the whole trajectory stays valid and therefore excludes episodes derailed by noise that answer correctly by chance. The benchmark is out of distribution for us: our perturbations are summarized entirely from large-scale real tool-call failures, were designed without any reference to AgentNoiseBench, and are injected by a different process; no AgentNoiseBench environment is used in training.

\subsection{Training Task Pools}
Two training pools must not be confused with the evaluation sets. \textbf{The \rsd{} ecosystem} supplies the SFT data: 557 Smithery-anchored environments, each with a mined database, validated tools, executable verifiers, and 60 tasks (Section~\ref{app:impl-stage1}), from which the \resft{} corpus is synthesized. \textbf{EnvScaler tasks} supply the RL data: ORPO directly reuses 2.5K EnvScaler RL tasks~\citep{song2026envscaler} rather than \rsd{}-synthesized tasks, and every algorithm in Figure~\ref{fig:algo-compare} trains on this same pool, isolating the objective from the Stage 1--2 data.

\section{Baselines}
\label{app:baselines}
Foundation-model rows are reference points at a different scale; bold and underline in Table~\ref{tab:main-results} are computed among 7B--8B methods only. Most baseline results are taken from the numbers reported in the corresponding papers. Only for methods that are not open-sourced, or for benchmarks that a paper does not report, do we reproduce the method locally, following the settings and technical route of the original paper.

\subsection{Foundation Models}
\textbf{GPT-5.2}, \textbf{Claude-4.5-Sonnet}, and \textbf{Gemini 3 Pro} are the proprietary reference points. \textbf{Qwen3-235B-A22B} is a large open-weight mixture-of-experts model included to separate scale from recipe. \textbf{Qwen3-8B} is the backbone for every trained method here, including \method{}, and is the control that matters: any margin over it is attributable to the training recipe rather than to capacity.

\subsection{Reflection Methods}
\begin{itemize}[leftmargin=1em,itemsep=2pt]
\item \textbf{Reflexion}~\citep{shinn2023reflexion} stores verbal self-reflection in an episodic buffer consulted on later attempts. It does not update parameters, so it bounds what inference-time reflection recovers.
\item \textbf{ExpeL}~\citep{zhao2024expel} accumulates cross-task experience as natural-language insights and retrieves similar past trajectories at test time: parameter-free, but with cross-task rather than within-episode memory.
\item \textbf{Agent-R}~\citep{yuan2025agentr} builds recovery trajectories by combining MCTS with model-guided selection of the revision point and learns timely correction by iterative self-training. It releases neither an 8B checkpoint nor its training data and does not report our benchmarks, so the rows marked $^*$ are our local reproduction following the original setting.
\item \textbf{NoisyAgent}~\citep{chen2026noisyagent} trains on clean/perturbed mixtures with adaptively increasing noise difficulty. Since no checkpoint is released, its row in Table~\ref{tab:main-results} (marked $^*$) is our local reproduction following the original setting, and the optimization algorithm compared in Figure~\ref{fig:algo-compare} is likewise reproduced from the description in the paper. Its row in Table~\ref{tab:noise-results} is quoted directly from the paper, whose reported $\tau^2$-Bench and VitaBench numbers come from RL on the evaluated domains themselves.
\end{itemize}

\subsection{Environment-Scaling Methods}
\begin{itemize}[leftmargin=1em,itemsep=2pt]
\item \textbf{Simulator}~\citep{li2025simia} is the model-simulated-environment route and \textbf{TOUCAN}~\citep{xu2025toucan} the trajectory-synthesis route.
\item \textbf{EnvScaler}~\citep{song2026envscaler} constructs executable environments with verifiable outcomes and is the strongest environment-scaling baseline in our tables; because the ORPO stage of \method{} reuses its RL task pool, it is the reference for attributing our margin to reflective supervision rather than coverage.
\item \textbf{AWM}~\citep{wang2026awm} learns an agent world model over transitions.
\item \textbf{EnvACE}~\citep{xu2026envace} internalizes dynamics through world rehearsal.
\item \textbf{Agent-World}~\citep{dong2026agentworld} synthesizes database-grounded environments of controllable difficulty over 2{,}000 environments.
\end{itemize}

\subsection{Agent Policy Optimization Algorithms}
These are the comparands in Figure~\ref{fig:algo-compare}, all trained from Qwen3-8B on the same 2.5K EnvScaler tasks under matched rollout, tool-call, and token budgets. \begin{itemize}[leftmargin=1em,itemsep=2pt]
\item \textbf{GRPO}~\citep{shao2024deepseekmath} normalizes verifier rewards within a rollout group and is the base of our own outcome term.
\item \textbf{DAPO}~\citep{yu2025dapo} adds decoupled clipping, dynamic sampling, token-level loss, and overlong-reward shaping.
\item \textbf{REINFORCE++}~\citep{hu2025reinforce++} strengthens critic-free optimization with baseline-based variance reduction.
\item \textbf{NoisyAgent RL}~\citep{chen2026noisyagent} appears here in its optimization role, as clean/noisy rollout mixing with an adaptive noise schedule, reproduced from the paper.
\item \textbf{SEED}~\citep{wu2026seed} adds same-action on-policy distillation to outcome-based Agent RL and is the most important comparand for ORPO, since it shares the estimator and differs mainly in what supplies the privileged context.
\end{itemize}

\section{Implementation Details}
\label{app:implementation}

\subsection{Training Guideline}
\label{app:impl-guideline}
Our aim is to test whether reflection can be mined, supervised, and internalized as a meta-skill, not to maximize a leaderboard number. All stages follow the cold-start-then-RL paradigm~\citep{dong2025toolstar} so RL does not begin from a policy that cannot complete a trajectory, and every stage and every trained method shares the \textbf{Qwen3-8B} backbone, with supervised fine-tuning implemented in LLaMA-Factory~\citep{zheng2024llamafactory} and reinforcement learning in VERL~\citep{sheng2025verl}. All RL baselines and ORPO use the same experimental settings and hyperparameters. Tokens belonging to tool returns and simulated user messages are excluded from the loss; only agent-generated tokens contribute gradient, which is the mask $\chi_{q,i,t,k}$ in Equations~\ref{eq:rl-hybrid} and~\ref{eq:opd}; the same exclusion applies to the \resft{} loss of Equation~\ref{eq:resft}.

\subsection{Stage 1: Environment Construction and Skill Discovery}
\label{app:impl-stage1}
\textbf{Environments.} Environment topics are 557 high-usage deployed MCP servers in the Smithery registry, ranked by the registry's cumulative \texttt{useCount}. Following Agent-World~\citep{dong2026agentworld}, a research agent with search, browser, code, and OS tools mines a topic-grounded database and expands it for $N=3$ complexification rounds; a coding agent then generates candidate tools with unit tests, and a tool is retained only if it compiles, passes its tests, respects state-transition constraints, and admits an executable verifier. Each environment contributes 60 verifier-backed tasks. Both agents are GPT-OSS-120B~\citep{openai2025gptoss}, matching the Agent-World setting so that environment construction is not advantaged by a stronger builder.

\textbf{Failure mining and injection.} Running agents across the nine public benchmarks listed below yields 177{,}700 tool returns of which 24{,}510 are errors. An analyzer model groups them by surface form into the eight families of Table~\ref{tab:noise-taxonomy}; the families are multi-label, so one return may land in several counts. During construction each tool-return position is perturbed independently with probability $p_{\mathrm{inj}}=0.2$ per Equation~\ref{eq:training-injection}. Both draws are balanced uniform: the family uniformly over the eight rather than in proportion to mined frequencies, and the realization uniformly among those compatible with the invoked tool. A clean task is kept only if it admits a verifiable solution, and a perturbed task only if it admits a verifiable recovery.

\textbf{Benchmarks replayed for failure mining.}
\label{app:mining-benchmarks}
The error returns above come from replaying agents across nine public agentic benchmarks, chosen to span interface types and task horizons rather than to maximize count. Many of them interact through MCP servers or shared workspaces and evaluate agents in those settings, which brings them close to real deployment. The nine benchmarks are MCP-Atlas~\citep{scaleai2026mcpatlas}, $\tau^3$-Bench~\citep{sierra2026tau3}, AutomationBench~\citep{zapier2026automationbench}, OfficeQA Pro~\citep{databricks2026officeqa}, JobBench~\citep{li2026jobbench}, Harvey LAB~\citep{harvey2026lab}, GDPval~\citep{openai2025gdpval}, Workspace-Bench~\citep{tang2026workspacebench}, and SkillsBench~\citep{benchflow2026skillsbench}. Together they reach MCP servers, REST suites, retrieval backends, and file or OS toolchains. They serve \emph{only} as a source of observed error returns for the taxonomy in Table~\ref{tab:noise-taxonomy}; none is used for training or reported as a result. 

\textbf{Sampling, pairing, and extraction.} Trajectories come from the nine-model pool of Table~\ref{tab:model-pool} under an identical tool-call limit, response budget, and sampling protocol, so differences reflect behavior rather than budget. DeepSeek-V4-Flash serves as both critique and analyzer model; together with the environment verifier it assigns the correctness label, and disagreements are discarded. Pairs form only within the same task, with at least five distinct tasks per environment; the analyzer then compares each pair and emits the environment record $\zeta_m$, from which 1{,}170 reflection meta-skills are summarized in total.

\textbf{Training-Trajectory Sources and Policy Initialization.} The diverse pool produces behavioral contrasts for skill mining only; it never supplies a training target. Every trajectory the policy is trained on comes from Qwen3-8B itself: $\mathcal{D}_{\mathrm{correct}}$ keeps only Qwen3-8B's verified successes, and all Stage-2 continuations are rolled out by the same backbone. A stronger model appears solely as the analyzer for error diagnosis and skill extraction, a task Qwen3-8B is not reliable enough to perform. At the level of the policy that generates trajectories we therefore stay at 8B throughout, which is what keeps the comparison against the 7B--8B baselines fair. 

We release the ReSFT and ORPO training data, the complete reflection skill
library, the Agent-Reflex-8B checkpoint, and the training and evaluation code
at \url{https://github.com/RUC-NLPIR/Agent-Reflex}.

\textbf{Skill record schema.} A skill is a structured record rather than free prose: a trigger over visible history with its supporting and contradicting evidence, a diagnosis naming the failure family, a recovery template stating when to retry, switch endpoint, ask the user, or abort, and reversibility and partial-effect flags. Provenance and replay-validation fields travel with each record so that a skill remains auditable.

\begin{table*}[t]
\centering
\footnotesize
\caption{\textbf{Empirical tool-failure taxonomy used for noise injection.}
Share is the fraction of non-exclusive pattern hits over 24,510 candidate error
returns, matching Figure~\ref{fig:motivation}(d). Families differ in whether
an unchanged retry is admissible at all.}
\label{tab:noise-taxonomy}
\setlength{\tabcolsep}{5pt}
\renewcommand{\arraystretch}{1.12}

\begin{tabularx}{\textwidth}{%
>{\raggedright\arraybackslash}p{3.0cm}%
>{\centering\arraybackslash}p{1.1cm}%
>{\raggedright\arraybackslash}p{3.25cm}%
>{\raggedright\arraybackslash}X}

\rowcolor{capsuleblue}
\color{white}\textbf{Failure family} &
\color{white}\textbf{Share} &
\color{white}\textbf{Typical signals} &
\color{white}\textbf{Recovery implication} \\

\midrule

\rowcolor{capsulelight}
Server/\allowbreak gateway unavailable &
33.6\% &
500/\allowbreak 502/\allowbreak 503/\allowbreak 504 &
Retry with bounded backoff; switch endpoint after the stop rule. \\

Authentication/\allowbreak permission &
17.9\% &
401/403, permission denied &
Refresh credentials, request authorization, or escalate; unchanged retry is insufficient. \\

\rowcolor{capsulelight}
Timeout &
14.7\% &
timeout, read timeout &
Retry or narrow the request after checking whether state changed. \\

Resource missing &
10.5\% &
404, not found &
Verify the identifier or discover the resource through search/list tools. \\

\rowcolor{capsulelight}
Argument/\allowbreak validation &
10.2\% &
400/422, invalid argument &
Repair schema, types, required fields, or enums before retrying. \\

Generic execution exception &
8.3\% &
tool/\allowbreak operation failed &
Inspect returned evidence and choose a typed recovery rather than blind repetition. \\

\rowcolor{capsulelight}
Rate limit &
4.5\% &
429, too many requests &
Back off, respect quota, batch requests, or switch an allowed provider. \\

Network connection &
0.3\% &
reset/\allowbreak refused/\allowbreak network error &
Treat as transient infrastructure failure unless repeated evidence suggests otherwise. \\

\bottomrule
\end{tabularx}
\end{table*}

\begin{table*}[t]
\centering
\footnotesize
\caption{\textbf{LLM agent pool for \rsd{} trajectory sampling.}
The pool only supplies behavioral contrasts for skill mining; all training
trajectories come from Qwen3-8B.}
\label{tab:model-pool}
\setlength{\tabcolsep}{7pt}
\renewcommand{\arraystretch}{1.18}

\begin{tabularx}{\textwidth}{
>{\raggedright\arraybackslash}p{3.2cm}
>{\raggedright\arraybackslash}p{4.2cm}
>{\raggedright\arraybackslash}X}

\rowcolor{capsuleblue}
\color{white}\textbf{Provider} &
\color{white}\textbf{Model family} &
\color{white}\textbf{Checkpoints for trajectory sampling} \\

\midrule

\rowcolor{capsulelight}
Alibaba &
Qwen3~\citep{yang2025qwen3} &
4B-Thinking, 8B, 14B, 32B, 235B-A22B \\

Meta &
Llama~\citep{grattafiori2024llama} &
3.1-8B-Instruct \\

\rowcolor{capsulelight}
Google &
Gemma~\citep{gemmateam2025gemma} &
4-31B-it \\

OpenAI &
GPT-OSS~\citep{openai2025gptoss} &
120B \\

\rowcolor{capsulelight}
DeepSeek-AI &
DeepSeek &
V4-Flash (284B) \\

\bottomrule
\end{tabularx}
\end{table*}

\subsection{Stage 2: Reflective Supervised Fine-Tuning}
\label{app:impl-stage2}
The three sampling strategies of Section~\ref{sec:resft} share one actor for continuation rollouts, $\pi_0=$ Qwen3-8B, the same backbone that is being trained, and one user simulator for UR, $U=$ GPT-4.1, following the $\tau^2$-Bench setting~\citep{barres2025tau2}. A sample enters the dataset only if the environment verifier accepts the resulting episode. The final corpus contains 54K trajectories, combining verified Qwen3-8B successes with reflective-correction samples spread over SR, KR, and UR. Two kinds of privileged information are kept out of every stored input. The simulator state snapshot is used to \emph{restore} the environment before a rollout resumes and never appears as tokens in the context, so the policy sees only what it would see at deployment. The skill text is removed from the stored input in SR and KR, and in UR it never enters the context at all, since the simulator rewrites it into a natural user message $u_{\skill}$; those synthesized user messages are retained, because a user turn is visible at deployment whereas a skill prompt is not. The UR simulator is prompted to cover direct correction, partial hints, ambiguity, and deliberately incorrect advice, so compliance is not always the right response.

\subsection{Stage 3: On-Policy Reflective Policy Optimization}
\label{app:impl-stage3}
ORPO starts from the ReSFT checkpoint with the skill library and reference policy frozen. ORPO directly reuses 2.5K EnvScaler RL tasks rather than \rsd{}-synthesized data, which keeps the comparison against EnvScaler and the other RL algorithms direct. For the OPD hyperparameters, we follow the standard settings of SEED~\citep{wu2026seed} to ensure a fair comparison. Each update samples a task group, rolls out $G=8$ ordinary trajectories of which a fraction $\nu=0.25$ runs in perturbed environments, normalizes advantages over the whole mixed group, and re-scores the recorded action tokens under the skill-augmented context. The skill branch is a forward pass only: it never queries the environment and its gradient is detached, so the marginal cost over plain Agent RL is a forward pass rather than a rollout. The confidence weight $\gamma$ is a soft token weight: it emphasizes tokens with a positive skill-induced shift and down-weights those whose shift is weak or negative, while every trajectory still contributes through the outcome term. We further conduct an ablation study on the perturbation fraction \(\nu\). Across the tested values, the performance remains relatively stable on the evaluated benchmarks, with only minor fluctuations and no consistent advantage for alternative settings. We therefore empirically set \(\nu=0.25\).

\paragraph{Meta-skill selection and environment alignment.}
The EnvScaler RL task pool is entirely distinct from the RSD task pool, but
the two are aligned at the environment level: the more than 100
environments used to construct EnvScaler's RL tasks are covered by the 557
high-usage MCP environments collected in Stage 1. Each environment and
synthesized task is assigned a canonical environment identifier, allowing
direct lookup of the corresponding skill:
\[
\kappa_q = \mathrm{Lookup}(\mathrm{env\_id}(q), \mathcal{K}).
\]
Thus, skills mined from RSD trajectories are transferred from the
environment level to different RL tasks instantiated within the same
environment, without semantic retrieval or task-level overlap. This
directly tests the reusability and transferability of the proposed
environment-level skills. At inference time, the skill prompt is removed,
so the deployed policy does not require access to the skill library.

\subsection{Evaluation Protocol and Contamination Control}
\label{app:impl-eval}
Every benchmark is scored with its official metric under the standard MCP format and task templates, following the EnvScaler protocol~\citep{song2026envscaler}. VitaBench uses official Avg@4 with the benchmark average recomputed from per-domain scores. AgentNoiseBench uses Stability-Gated Success, and \method{} is measured under exactly that protocol. Following EnvScaler and Agent-World~\citep{song2026envscaler,dong2026agentworld}, every score we measure is the mean of three independent runs, and decoding settings, maximum turns, and tool-call limits follow each benchmark's standard setting. Most baseline numbers are taken from the results reported in the corresponding papers; only for methods that are not open-sourced or benchmarks that a paper does not report do we reproduce locally, following the original settings and technical route.

\textbf{Noise taxonomy and row protocol in Table~\ref{tab:noise-results}.} \textit{Tool Noise} aggregates fault, incomplete, induce, redundant and wrong returns; \textit{User Noise} aggregates ambiguity, topic shift, conflict, redundancy and boundary probing. The \textit{Clean} results of foundation models follow the original AgentNoiseBench report and may therefore differ slightly from Table~\ref{tab:main-results}. $^*$Reproduced locally because the paper releases no 8B checkpoint or training data. The NoisyAgent row is quoted from its paper, which reports only per-domain aggregates, so its two noise buckets are left blank.

Because agent training on synthesized environments is easy to contaminate, we state the controls explicitly. Splits are preserved at the \emph{environment-family} level rather than the task level, since tasks within one environment share a database and tool set; generated instructions and skill text are checked for near-duplicates against benchmark instructions and tool schemas; and evaluation reference answers never appear in any analyzer, teacher, or user-simulator prompt. We release the task identifiers of all training data and the tool-fault injector used during training.

\section{Supplement Experimental Results}
\label{app:supp-results}
This section reports the fine-grained noise attribution deferred from Section~\ref{sec:experiments}.

\subsection{Does \method{} Act Reflectively Under Noise?}
\label{app:reflective}
\begin{figure}[t]
    \centering
    \includegraphics[width=\linewidth]{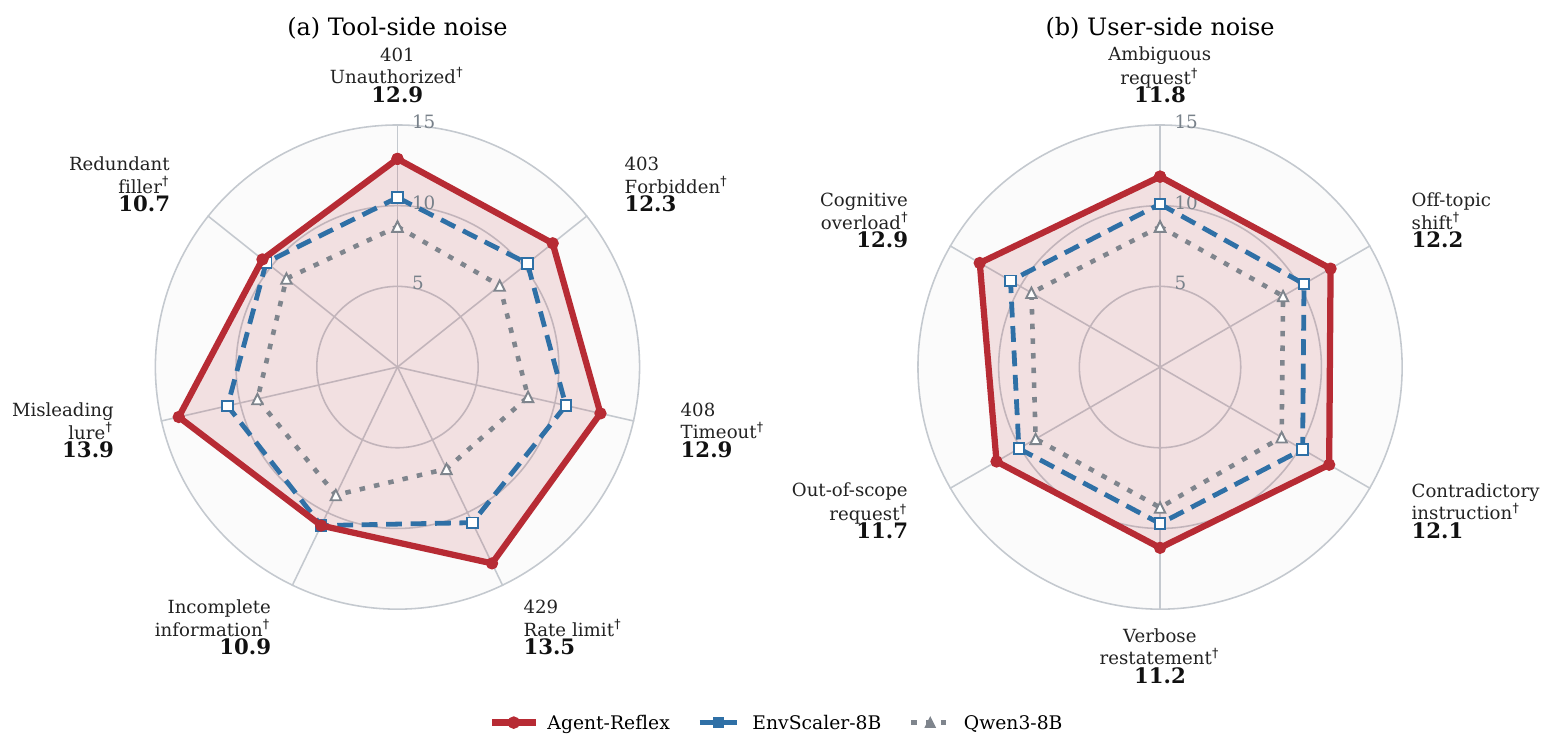}
\caption{\textbf{Per-axis Stability-Gated Success on AgentNoiseBench's native taxonomy}, normalized by the three-model mean on each axis (10 = that average). \textbf{(a)} Seven tool-side axes. \textbf{(b)} Six user-side. $^*$ marks a lead clearing a 95\% paired-bootstrap interval against \emph{both} baselines, $^\dagger$ the backbone alone.}
    \label{fig:noise-radar}
\end{figure}

Task success alone cannot show whether an agent repaired for the right reason. We therefore split the AgentNoiseBench noise settings into finer diagnostic slices (seven tool-side and six user-side axes, which Table~\ref{tab:noise-results} aggregates into its tool- and user-noise buckets) and score each with the benchmark's Stability-Gated Success ($\mathrm{SGA}=I_{\mathrm{traj}}\!\times\!I_{\mathrm{task}}$), which credits a rollout only if its trajectory was never derailed by the injected artifact, discarding turns pulled off course that answer correctly by luck. Each axis is normalized by its own three-model mean so noise-family difficulty cannot dominate the shape, with 95\% paired bootstrap intervals over 165 tasks per axis. As shown in Figure~\ref{fig:noise-radar}, \method{} leads on all thirteen axes, with the margin widest exactly where each part of the design is meant to act.

\textbf{Tool side.} The two axes clearing the interval against \emph{both} baselines are misleading lure (13.9, the largest tool-side margin) and incomplete information (10.9), the families whose returns are well-formed, so no schema check fires and only the evidence inside the payload reveals the fault. This is what \rsd{} contributes and environment scaling cannot: more clean environments add successful trajectories, never a record of which well-formed return silently lied. On the four hard-failure templates, where the status code already names the fault, gains over the backbone are uniform (12.3--13.5).

\textbf{User side.} The largest margin is cognitive overload (12.9), the only user-side axis also clearing the interval against the stronger EnvScaler-8B, and the one \resft{}'s user-reflection supervision targets directly: when the instruction itself is the defect, the repair is to re-scope the request rather than re-plan the tools, a behavior no tool-side reward can teach. Pooled over both sides, \method{} is derailed on 28.5\% of rollouts against 44.8\% for EnvScaler-8B and 56.1\% for the backbone. Because the metric gates on trajectory validity, none of this is reachable by recovering an answer after being derailed: the gains are repair, not tolerance.

\subsection{Why AgentNoiseBench Is Out of Distribution for \method{}}
\label{app:ood}
AgentNoiseBench is never consulted while \rsd{} is built: neither its taxonomy nor its injector informs our perturbations, and the two noise designs are constructed in entirely different ways. The noise used by \method{} comes entirely from summarizing large-scale real tool-call failures: the eight families of Table~\ref{tab:noise-taxonomy} are counted from 24{,}510 error returns observed while replaying agents on nine public agentic benchmarks, and are injected into tool returns of our own synthesized environments during construction and training (Equation~\ref{eq:training-injection}). AgentNoiseBench instead derives its categories from its own empirical analysis of interaction noise and uses a separate injector that rewrites tool returns and user utterances of the $\tau^2$-Bench and VitaBench environments at evaluation time, covering seven tool-side and six user-side diagnostic axes. The two therefore differ in their source of noise, in the environments they act on, and in the injection process, and no AgentNoiseBench environment, task, or injector is used anywhere in training. Performance here therefore measures whether a reflective policy transfers to unseen perturbations, not whether a seen noise distribution is recalled.

\subsection{Representative Environment-Specific Skills}
\label{app:skill-examples}
To make the mined library concrete, we show five records from substantially different domains in \texttt{environment\_skills.jsonl}. Each capsule preserves the four fields used by \rsd{}: environment, representative tools, bad pattern, and extracted skill.

\begin{skillcapsule}{AKShare One MCP Server}
\skillblock{blockblue}{Environment}{Market-data and instrument-registry service. It exposes historical and real-time quotes, news, financial statements, insider-trading records, and instrument metadata, with date, source, pagination, and symbol constraints.}

\skillblock{blockblue}{Representative tools}{\begin{itemize}[leftmargin=1.2em,itemsep=0pt,topsep=1pt]
  \item \texttt{resolve\_symbol}: map a ticker to a normalized instrument record.
  \item \texttt{list\_instruments}: filter and paginate the instrument registry.
  \item \texttt{get\_hist\_data}: return OHLCV bars with optional adjustment and indicators.
  \item \texttt{get\_realtime\_data}: return the latest quote snapshot.
  \item \texttt{get\_balance\_sheet} / \texttt{get\_cash\_flow}: return statement rows.
\end{itemize}}

\skillblock{blockred}{Bad patterns}{\begin{itemize}[leftmargin=1.2em,itemsep=0pt,topsep=1pt]
  \item Fail to verify the target instrument and mix similar objects.
  \item Ignore the requested date, source, count, or field boundary and report an incomplete result as exhaustive.
\end{itemize}}

\skillblock{blockgreen}{Extracted skills}{\begin{itemize}[leftmargin=1.2em,itemsep=0pt,topsep=1pt]
  \item Confirm the resource, object, and requested filter boundary before querying.
  \item Reconcile every returned row against symbol, range, count, and field types; defaults, summaries, and accepted requests are not evidence of returned data.
\end{itemize}}
\end{skillcapsule}

\begin{skillcapsule}{DeepResearch}
\skillblock{blockblue}{Environment}{Stateful research workflow: a query creates a session, iterative steps add notes and citations, and a final report is available only after the session reaches a reportable state. Session, step, and source records are separately addressable and may be paginated.}

\skillblock{blockblue}{Representative tools}{\begin{itemize}[leftmargin=1.2em,itemsep=0pt,topsep=1pt]
  \item \texttt{initialize-research}: create a session with a query and target depth.
  \item \texttt{execute-research-step}: advance exactly one research step.
  \item \texttt{get-research-session} / \texttt{list-research-steps}: expose status and progress.
  \item \texttt{list-sources}: enumerate discovered sources.
  \item \texttt{generate-report}: compile a report only from a reportable session.
\end{itemize}}

\skillblock{blockred}{Bad patterns}{\begin{itemize}[leftmargin=1.2em,itemsep=0pt,topsep=1pt]
  \item Continue reasoning after a key return is missing or a call fails.
  \item Treat a default parameter or an old session snapshot as the current state, and confuse session, step, or report identifiers.
\end{itemize}}

\skillblock{blockgreen}{Extracted skills}{\begin{itemize}[leftmargin=1.2em,itemsep=0pt,topsep=1pt]
  \item Map each input condition to the required target field and execute the workflow in state order.
  \item Re-check session, source, and report records after every transition; never fill missing values by guesswork or conflate stage identifiers.
\end{itemize}}
\end{skillcapsule}

\begin{skillcapsule}{Airtable Server}
\skillblock{blockblue}{Environment}{Structured-table environment with nested bases, tables, fields, views, and records. Operations use opaque identifiers, cursor pagination, schema constraints, soft deletion, and authorization checks.}

\skillblock{blockblue}{Representative tools}{\begin{itemize}[leftmargin=1.2em,itemsep=0pt,topsep=1pt]
  \item \texttt{list\_bases}: discover visible bases.
  \item \texttt{list\_tables}: enumerate tables within a selected base.
  \item \texttt{list\_fields}: verify the table schema.
  \item \texttt{search\_records} / \texttt{list\_records}: return filtered or paginated records.
  \item \texttt{get\_record}: read one record by its table-scoped identifier.
  \item \texttt{update\_record} / \texttt{delete\_record}: mutate records under schema and lifecycle constraints.
\end{itemize}}

\skillblock{blockred}{Bad patterns}{\begin{itemize}[leftmargin=1.2em,itemsep=0pt,topsep=1pt]
  \item Act on a similar search hit without confirming the target base and table.
  \item Omit the record identifier or post-write read, so the operation can affect the wrong scope and remain unverifiable.
\end{itemize}}

\skillblock{blockgreen}{Extracted skills}{\begin{itemize}[leftmargin=1.2em,itemsep=0pt,topsep=1pt]
  \item Resolve base, table, and record identifiers in that order and compare key fields with the user's constraints.
  \item Update only after an exact match, then re-read the record to verify committed values and lifecycle state.
\end{itemize}}
\end{skillcapsule}

\begin{skillcapsule}{Hotel Booking Server}
\skillblock{blockblue}{Environment}{Stateful hotel-search and booking workflow. A location query creates a search session, hotel results are paginated within that session, rate details are scoped to a returned hotel, and a booking attempt produces a lifecycle record and payment link.}

\skillblock{blockblue}{Representative tools}{\begin{itemize}[leftmargin=1.2em,itemsep=0pt,topsep=1pt]
  \item \texttt{find-place}: resolve free-text locations to coordinates.
  \item \texttt{search-hotels}: start a session with dates, guests, and filters.
  \item \texttt{load-more-hotels}: advance the session cursor.
  \item \texttt{get-search-session}: verify query, pagination, and lifecycle state.
  \item \texttt{get-hotel-details}: expand a hotel returned by the session.
  \item \texttt{book-hotel}: initiate a booking for a currently valid rate.
\end{itemize}}

\skillblock{blockred}{Bad patterns}{\begin{itemize}[leftmargin=1.2em,itemsep=0pt,topsep=1pt]
  \item Mix hotels or rates from different search sessions, or treat a paginated summary as complete.
  \item Initiate booking without checking current availability and whether the rate belongs to the selected hotel.
\end{itemize}}

\skillblock{blockgreen}{Extracted skills}{\begin{itemize}[leftmargin=1.2em,itemsep=0pt,topsep=1pt]
  \item Preserve the session identifier and verify query parameters, pagination, hotel identity, and room/rate constraints at each step.
  \item Re-read the returned hotel/rate and booking records; treat payment-link creation as an initiated booking, not completed payment.
\end{itemize}}
\end{skillcapsule}

\begin{skillcapsule}{Semantic Scholar Server}
\skillblock{blockblue}{Environment}{Academic-literature environment for searching papers and authors and traversing citation/reference edges. Search is truncated, while detailed metadata and citation edges require stable paper or author identifiers and are cached for inspection.}

\skillblock{blockblue}{Representative tools}{\begin{itemize}[leftmargin=1.2em,itemsep=0pt,topsep=1pt]
  \item \texttt{search\_semantic\_scholar}: find candidate papers.
  \item \texttt{get\_semantic\_scholar\_paper\_details}: resolve a full paper record.
  \item \texttt{get\_semantic\_scholar\_author\_details}: resolve a full author record.
  \item \texttt{get\_semantic\_scholar\_citations\_and\_references}: retrieve verified citation edges.
  \item \texttt{list\_semantic\_scholar\_papers} / \texttt{get\_semantic\_scholar\_search\_query}: audit cached records and query status.
\end{itemize}}

\skillblock{blockred}{Bad patterns}{\begin{itemize}[leftmargin=1.2em,itemsep=0pt,topsep=1pt]
  \item Fail to keep a unique paper or author identifier attached to the retrieved record, then use a similar or truncated hit for citation statistics.
  \item Treat a missing result or service error as confirmed metadata, leaving the final claim without verifiable evidence.
\end{itemize}}

\skillblock{blockgreen}{Extracted skills}{\begin{itemize}[leftmargin=1.2em,itemsep=0pt,topsep=1pt]
  \item Verify title, authors, year, DOI, and stable identifier before reading citation edges; retrieve the target record in a second step.
  \item Cite only fields present in the returned record; if the target is missing or the service fails, mark the claim unconfirmed and refine the query.
\end{itemize}}
\end{skillcapsule}

\section{Limitations}

Our reflection skill library is mined offline from a fixed snapshot of a large collection of executable environments. Consequently, it does not automatically adapt to changes in tool interfaces, service behavior, or environment dynamics. Such changes may require periodic re-mining or lightweight maintenance to keep the library aligned with the current environments.

Online updating, however, does not necessarily yield broadly generalizable skills. When the available environments or tasks lack sufficient diversity, skills summarized online may remain specific to individual environments or local failure patterns. The broader environmental coverage used for offline skill discovery may therefore help explain the stronger generalization of \method{} relative to SEED, although isolating this factor would require further controlled experiments. This highlights the importance of balancing update frequency with sufficient environmental coverage for skill generalization.

Future work will investigate online skill summarization and incremental updating across larger and more diverse environment collections, allowing the library to incorporate new environments and emerging failure patterns while supporting cross-environment generalization.

\section{Broader Impacts and Future Directions}
\label{app:limitations}
\textbf{Scope of the current recipe.} \method{} is validated on 557 executable environments anchored in live MCP services, eight empirically mined tool-failure families, and six evaluation suites spanning long-horizon tool use and noisy interaction, so the recipe is exercised well beyond a single benchmark. Its three stages are also deliberately modular: the skill library, the reflective supervision, and the token-level objective can each be replaced without touching the others, which makes the extensions below incremental rather than redesigns.

\textbf{Future directions.} Three extensions follow naturally. First, the failure taxonomy is textual, mined from what an agent \emph{sees} when a service fails; extending it to temporal signatures such as latency, concurrency, and partial writes would widen the range of deployment faults a reflective policy can anticipate. Second, the library is mined once at 8B scale, and testing whether it transfers across capacities, and whether skills compose across environments rather than only within them, would turn it from a per-recipe artifact into a reusable resource. Third, the privileged slot in ORPO accepts any environment-level artifact, so targets richer than reflection rules, verified sub-plans or tool contracts for instance, can be distilled with exactly the same estimator.

\textbf{Broader impacts.} An agent that recovers from failure is more useful and also more consequential when wrong, since it takes more actions in the world than one that halts. Our skill records therefore carry explicit reversibility and partial-effect fields, and recovery templates distinguish asking the user from acting unilaterally. The \resft{} user-reflection data deliberately includes incorrect advice so the policy learns evidence-sensitive replanning rather than obedience. All environments are synthesized from public service specifications and contain no real user data, and because training perturbations are mined from real error logs, we release failure counts and families rather than per-service identities.

\section{Ethics statement}

This work does not involve any issues related to the Code of Ethics.

\end{document}